\documentclass[10pt]{article}
\usepackage[T1]{fontenc}
\usepackage[utf8]{inputenc}
\usepackage{mathptmx}
\usepackage{courier}
\usepackage[scaled=0.9]{helvet}
\usepackage[margin=1in]{geometry}
\usepackage{microtype}
\usepackage{graphicx}
\usepackage{booktabs}
\usepackage{longtable}
\usepackage{array}
\usepackage{calc}
\usepackage{etoolbox}
\usepackage{caption}
\usepackage{url}
\usepackage[hidelinks]{hyperref}
\usepackage{enumitem}
\setlist{itemsep=2pt,topsep=3pt}
\AtBeginEnvironment{longtable}{\footnotesize}
\providecommand{\tightlist}{\setlength{\itemsep}{0pt}\setlength{\parskip}{0pt}}
\makeatletter
\def\maxwidth{\ifdim\Gin@nat@width>\linewidth\linewidth\else\Gin@nat@width\fi}
\makeatother
\setkeys{Gin}{width=\maxwidth,keepaspectratio}

\title{Where Does an On-Device Prescription Reader Lose
Medications? A Frozen-Replay Study, from a 270M Language
Model to a 2B Vision-Language Model}
\author{Ilan Sezhiyan
Jayaraman\thanks{Written in an individual capacity. The views and work are the author's own and are not made on behalf of any employer. Contact: jayaramanilan@zohomail.in. Version 1 of this work, which covers only the 29 October pipeline, is archived on Zenodo, doi:10.5281/zenodo.23795680.}}
\date{3 November 2026}
\begin{document}
\maketitle
\begin{abstract}
\noindent On-device models let a phone read a prescription
without sending the photograph anywhere, but small models
fail in ways that are hard to attribute: to the camera, the
OCR, the model, or the rules around it. We study a
phone-based pipeline that reads a prescription photograph
with PaddleOCR, extracts medications with a 270M-parameter
language model, and filters the result with a deterministic
backstop before showing suggestions. Replaying identical
frozen inputs through each configuration, with deterministic
generation verified, we compare configurations with paired
document-level intervals on 49 captures of 15 synthetic
prescriptions (6 typed, 9 handwritten by one writer).
Production recall is 8\%, and 0 of 45 on handwriting. The
backstop removes no correct medication and cuts unsafe
outputs from 45 to 1. On handwriting nearly all medications
never reach the model in readable form; on typed sheets the
model mostly picks the wrong text. We trace much of the
handwriting floor to a defect in our own crop step, then
replace the pipeline: rotated-box crops, a 2B-parameter
vision-language model that re-reads prescription lines and
extracts the medications, and per-line numeric grounding. On
the same captures recall rises from 8\% to 67\% (paired +59
points, 95\% CI +40 to +78): typed 19\% to 100\%,
handwritten 0 to 20 of 45, with no unsafe output. Rotated
crops account for most of the reading gain and the model for
most of the recall gain, and the OCR confidence gate is now
the main limit. The cost is 61 s per photograph against
about 12 s, and about 1.9 GB of memory. An earlier claim
that better OCR lowered recall does not survive the
controlled comparison, and we withdraw it. All results are
in-sample, from one writer and one device.
\end{abstract}
\hypertarget{introduction}{%
\section{1 Introduction}\label{introduction}}

Handwritten and printed prescriptions are a daily source of
structured information that people must read correctly:
which drug, how often, for how long. Reading them by machine
is attractive for the person holding the paper, who may want
a reminder, and sensitive for the same reason: the
photograph names a patient, a diagnosis and a medication.
Running the whole pipeline on the phone avoids sending that
photograph to a server. It also restricts the model. A
language model that fits a phone has a few hundred million
parameters, runs at a few tokens per second, and is easy to
distract with text on the page that is not a medication.

Whether such a pipeline works is usually answered with an
end-to-end accuracy figure. That figure hides where the
pipeline fails. A prescription passes through a camera, an
OCR engine, a quality check, a language model and a set of
rules, and a missed medication could have been lost at any
of them. Studies that vary one component on separate
captures also confound the comparison with the capture
itself: a different photograph, a different operator, a
different lighting. We encountered this directly: an early
result suggested that a better OCR engine lowered recall,
and it did not survive a controlled comparison.

This paper asks where an on-device prescription reader loses
medications and which common changes recover them. We hold
the input fixed by replaying frozen captures through each
configuration on the phone itself, verify that this is
faithful, and compare configurations with paired
document-level intervals. We first study the pipeline as it
stood on 29 October 2026: PaddleOCR PP-OCRv5, a
270M-parameter Gemma 3 model, and a deterministic
fabrication backstop, on 49 captures of 15 synthetic
prescriptions. We then report the pipeline that replaced it
on 3 November, built around a 2B-parameter vision-language
model, and measure how much of the change comes from which
part.

\textbf{Research questions.}

\begin{itemize}
\tightlist
\item
  \textbf{RQ1.} At which stage is recall lost, for typed and
  for handwritten input?
\item
  \textbf{RQ2.} Does a deterministic backstop keep unsafe
  output from the user, and what does it cost in recall?
\item
  \textbf{RQ3.} Which common mitigations (repetition
  penalty, line-scoped input, a larger model, numeric
  grounding) change recall or dosing accuracy on identical
  inputs?
\item
  \textbf{RQ4.} (Exploratory) Once an upstream defect is
  repaired, where does the bottleneck move?
\item
  \textbf{RQ5.} What does a document-level OCR confidence
  gate miss?
\item
  \textbf{RQ6.} (Post hoc) Does replacing the extractor with
  a vision-language model that re-reads the prescription
  lines recover handwriting, which part of the gain comes
  from the OCR fix and which from the model, and what does
  it cost on the phone?
\end{itemize}

\textbf{Findings.} Production recall is 8\% overall and 0 of
45 on handwriting. The backstop removed no correct
medication in any configuration and reduced unsafe outputs
from 45 to 1. On handwriting, 43 of 45 medication slots
never reached the model in readable form; on typed sheets,
22 of 25 misses were on the model side. A scoped prompt, a
switched-off repetition penalty and a larger model each
raised recall on typed text, and none helped handwriting.
Numeric grounding introduced 7 of the 8 dosing errors on
correctly named medications. We withdraw an earlier claim
that better OCR lowered recall. Exploratory analysis shows
that axis-aligned line crops in our OCR port read sloping
handwriting as blank, and that repairing them moves the
bottleneck to the gate, the extractor and name grounding.

Replacing the pipeline with rotated-box crops, a
2B-parameter vision-language model that re-reads and
extracts, and per-line numeric grounding raised recall on
the same captures from 8\% to 67\% (paired +59 points, 95\%
CI +40 to +78), typed recall from 19\% to 100\% and
handwritten recall from 0 of 45 to 20 of 45, with no unsafe
output. Rotated crops account for most of the gain in
reading accuracy and the model for most of the gain in
recall, and the OCR confidence gate is now the main limit.
The price is 61 s per photograph against about 12 s, and
about 1.9 GB of memory against about 545 MB.

\textbf{Contributions.}

\begin{enumerate}
\def\labelenumi{\arabic{enumi}.}
\tightlist
\item
  A frozen-replay method for comparing on-device
  configurations on identical inputs, with checks that
  generation is deterministic, that live output equals
  replay output, and that OCR is reproducible.
\item
  A stage-attributed failure taxonomy showing that the
  dominant loss differs between typed and handwritten input.
\item
  Evidence that a small deterministic backstop is precise
  (no correct medication removed) and effective (unsafe
  outputs 45 to 1).
\item
  Negative results on two common safeguards: a repetition
  penalty that costs recall, and numeric grounding whose
  fixed window crosses prescription lines.
\item
  A documented withdrawal of an earlier cross-run result,
  with intervals.
\item
  A case study of how one upstream defect hid every
  downstream comparison, and of where the bottleneck moves
  when it is repaired.
\item
  A test set of 15 synthetic prescriptions, with ground
  truth given in Appendix A and the scoring and attribution
  rules described in Section 4.6.
\item
  A before-and-after evaluation of a vision-language
  extractor on the same frozen captures, with an ablation
  that separates the effect of the OCR crop fix from the
  effect of the model's re-read and extraction (Section
  5.8).
\item
  The cost of that gain on a mid-range phone, reported
  alongside the accuracy.
\end{enumerate}

\textbf{Scope.} The evidence is in-sample: one writer, 15
synthetic documents, one phone, no held-out set (Section 7).
The results locate failures in one pipeline and measure one
replacement of it. They do not estimate performance on real
prescriptions, and they say nothing about clinical accuracy.

\textbf{Organisation.} Section 2 reviews related work.
Section 3 describes the pipeline and threat model, Section 4
the experimental design, Section 5 the results, including
the replacement pipeline in Section 5.8, Section 6 discusses
them, Section 7 lists the threats to validity, and Section 8
concludes.

\hypertarget{related-work}{%
\section{2 Related work}\label{related-work}}

\textbf{Prescription datasets and recognition.} Public
resources fall into two kinds. Word-level collections give
cropped images of single handwritten words: RxHandBD has
5,578 word images with a vocabulary of 1,559 entries
{[}Islam 2026{]}, and the Doctor's Handwritten Prescription
BD dataset has 4,680 word images over 78 medicine classes
{[}Rahim 2026{]}. These support recognition models but
contain no page layout, no surrounding text and no
extraction task. At the document level, MIRAGE provides
743,118 simulated prescription images from 1,133 doctors
across India and fine-tunes three multimodal language models
to extract medication names and dosages, reporting 82\%
accuracy {[}Mankash et al.~2024{]}. MIRAGE is closest to our
setting in task and in its use of simulated documents, but
it concerns large multimodal models trained for the task,
and reports one accuracy figure. We run a 270M-parameter
model on a phone and ask where in the pipeline each
medication is lost. Our test set is far smaller and serves a
different purpose: a controlled comparison, not training or
benchmarking.

\textbf{OCR engines.} Tesseract is a long-standing
open-source OCR engine {[}Smith 2007{]}. TrOCR pairs a
pre-trained image encoder with a pre-trained text decoder
and reports a character error rate of 2.89 on the IAM
handwriting dataset for its large model {[}Li et
al.~2021{]}. The PaddleOCR 3.0 report presents PP-OCRv5 as a
multilingual text recognition model with fewer than 100
million parameters, aimed at competing with much larger
vision-language models {[}Cui et al.~2025{]}. We use
PP-OCRv5 and Tesseract as installed engines, not as subjects
of study. Our contribution here is negative and specific:
our port of the PaddleOCR crop step contained a defect that
made OCR on sloping handwriting look far worse than the
engine's design suggests, and it was found only because
failures were attributed to stages.

\textbf{Extraction from OCR output and error propagation.}
Medication extraction from clean clinical text is well
studied; the 2018 n2c2 shared task reported top F1 scores of
0.94 for concept extraction on discharge summaries {[}Henry
et al.~2020{]}, but its input was typed text, not the output
of an OCR engine. For prescription images, RxLens describes
a deployed cloud pipeline that redacts personal data, runs
OCR, extracts medications with commercial language models
and matches them against a product catalogue, evaluated on
two proprietary datasets {[}Jagatap et al.~2025{]}. A recent
pipeline paper analyses how OCR errors propagate to
medication extraction and argues for joint evaluation of the
two stages, using public handwriting and clinical-text
benchmarks {[}Devi and Prakash 2026{]}; the abstract we read
reports no quantitative results. These works motivate
evaluating the stages together. They do not, as far as we
found, compare configurations of a small on-device model on
identical frozen inputs, attribute each missed medication to
the stage that lost it, or report documents rejected by a
quality gate as part of recall.

\textbf{Small models, constrained decoding and decoding
penalties.} We use the 270M-parameter model of the Gemma 3
family {[}Gemma Team 2025; Google 2025{]}, with Qwen3-0.6B
as the larger alternative {[}Yang et al.~2025{]}, run
through llama.cpp {[}ggml-org{]}, whose grammar support
constrains output to a fixed JSON structure;
grammar-constrained generation is also studied in systems
such as SynCode {[}Ugare et al.~2024{]}. The replacement
pipeline uses Qwen2-VL-2B-Instruct, a vision-language model
whose image encoder accepts images at their native
resolution {[}Wang et al.~2024{]}. Using such a model to
re-read lines of handwriting extends the engine-level
comparison of the preceding paragraph: it reads the line
image directly, where PaddleOCR's recogniser reads a crop. A
sampling penalty that discounts the scores of previously
generated tokens was introduced in CTRL {[}Keskar et
al.~2019{]}. Its effect on structured extraction, where the
correct answer is often a copy from the prompt, is rarely
tested. We find that the penalty costs recall (Section 5.4);
the explanation, that the prompt primes the penalty with the
drug names, remains a hypothesis.

\textbf{Rejection and grounding.} Rejecting uncertain inputs
is studied as selective classification, which trades
coverage against risk {[}Geifman and El-Yaniv 2017{]}. We
apply that framing to an OCR confidence gate and report
coverage, selective risk and false-reject rate (Section
5.5). Language models also state things the input does not
support, a problem surveyed under hallucination {[}Ji et
al.~2022{]}. Our backstop is a simple response: keep a
medication only if the OCR text supports it. What we add is
a measurement of its cost, and evidence that tolerating OCR
errors in the match also tolerates model errors (Section
6.4).

\textbf{Evaluation practice.} Comparing systems on separate
captures confounds them with the capture. We use frozen
inputs, deterministic generation and paired differences, and
resample documents, not captures, because the captures of
one document are correlated {[}Field and Welsh 2007; Efron
and Tibshirani 1993{]}.

\hypertarget{system-and-threat-model}{%
\section{3 System and threat
model}\label{system-and-threat-model}}

\hypertarget{context}{%
\subsection{3.1 Context}\label{context}}

The pipeline studied here is the \texttt{read\_prescription}
tool of AgentOS, an on-device Android layer in which a small
language model turns a user's request into calls on native
device capabilities instead of opening apps {[}Jayaraman
2026{]}. This paper does not evaluate AgentOS. We use the
prescription tool because it is a small, self-contained
example of a general problem: a language model with a few
hundred million parameters must turn noisy text from a
camera into structured data, on a phone, with no network,
and a wrong answer can matter.

\hypertarget{pipeline}{%
\subsection{3.2 Pipeline}\label{pipeline}}

A user photographs a prescription. Five stages follow
(Figure 1; Section 4.3 gives the settings):

\begin{enumerate}
\def\labelenumi{\arabic{enumi}.}
\tightlist
\item
  \textbf{OCR.} PaddleOCR PP-OCRv5 returns the recognised
  text and a document-level confidence.
\item
  \textbf{OCR gate.} If the confidence is below
  \ensuremath{\tau}, the capture is rejected and the user is
  asked to retake the photo. No extraction runs.
\item
  \textbf{Extraction.} Gemma 3 270M, quantised, runs on the
  device through llama.cpp. A fixed prompt with worked
  examples and a JSON grammar constrain the output to a list
  of medications, each with a name, dose, frequency and
  duration. A repetition penalty discourages the model from
  copying prompt text.
\item
  \textbf{Fabrication backstop.} Deterministic rules,
  described below, remove outputs that the OCR text does not
  support and repair truncated output.
\item
  \textbf{SUGGEST cards.} Each surviving medication becomes
  a card. The user can convert a card into a reminder or
  dismiss it. The pipeline never creates a reminder by
  itself.
\end{enumerate}

\begin{figure}
\centering
\includegraphics[width=1\textwidth,height=\textheight]{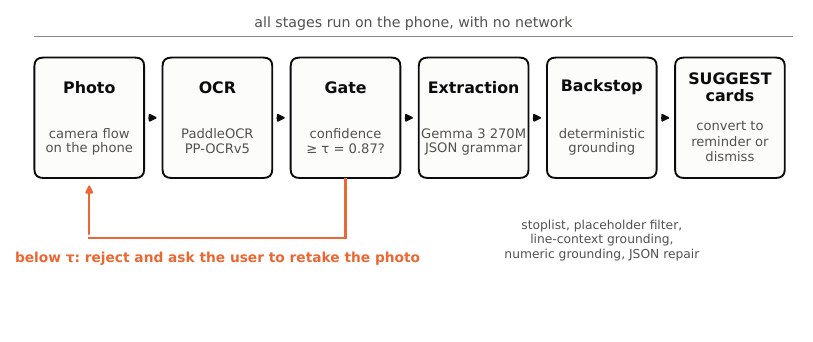}
\caption{Pipeline under test (the 29 October version; Table
11 gives the changes made on 3 November). A capture below
the confidence threshold \ensuremath{\tau} is rejected and
the user is asked to retake the photo. Otherwise the OCR
text goes to the extractor and then to the backstop, and
surviving medications become SUGGEST cards that the user
converts to reminders or dismisses.}
\end{figure}

\textbf{The backstop.} It has five parts, each aimed at an
observed failure:

\begin{longtable}[]{@{}
  >{\raggedright\arraybackslash}p{(\columnwidth - 4\tabcolsep) * \real{0.2045}}
  >{\raggedright\arraybackslash}p{(\columnwidth - 4\tabcolsep) * \real{0.4432}}
  >{\raggedright\arraybackslash}p{(\columnwidth - 4\tabcolsep) * \real{0.3523}}@{}}
\toprule\noalign{}
\begin{minipage}[b]{\linewidth}\raggedright
Component
\end{minipage} & \begin{minipage}[b]{\linewidth}\raggedright
What it does
\end{minipage} & \begin{minipage}[b]{\linewidth}\raggedright
Failure it targets
\end{minipage} \\
\midrule\noalign{}
\endhead
\bottomrule\noalign{}
\endlastfoot
Stoplist & Drops names that are known non-medications
(clinic words, form words, advice words) & The model copying
header, clinic or advice text as a drug \\
Placeholder filter & Drops names from the prompt's worked
examples & The model repeating an example drug (we saw
``ZolvaX'') \\
Line-context grounding & Keeps a name only if it appears in
the OCR text on a line that looks like a prescription line,
with a dose or dosage form & Names that appear elsewhere on
the page \\
Numeric grounding & Replaces the model's frequency and
duration with values found within 50 characters of the name
in the OCR text & Invented or misplaced numbers \\
JSON repair & Closes a truncated array so that complete
entries are kept & Output cut off by the token limit \\
\end{longtable}

The backstop is a set of rules, not a learned model. Its
design principle is \textbf{fail toward omission}: it is
better to propose nothing than to propose something the text
does not support.

\hypertarget{threat-model}{%
\subsection{3.3 Threat model}\label{threat-model}}

\textbf{What we protect.} The user's trust in a medication
card. A card that names a drug, dose or duration invites the
user to treat it as read from the prescription, and to turn
it into a reminder.

\textbf{Failure modes, ordered by consequence.} This
ordering is a design choice of the system, not a clinical
finding.

\begin{enumerate}
\def\labelenumi{\arabic{enumi}.}
\tightlist
\item
  \textbf{Fabricated medication.} A card names a drug that
  is not on the prescription (a clinic name, a signature, an
  example drug, a garbled word). We call this an
  \emph{unsafe output}.
\item
  \textbf{Wrong detail on a correct medication.} The name is
  right but the frequency or duration is wrong (a
  \emph{dosing error}).
\item
  \textbf{Omission.} A medication on the prescription yields
  no card. The user sees fewer cards than medications.
\item
  \textbf{Silent failure.} The pipeline proposes nothing and
  does not say why.
\end{enumerate}

Our design treats 1 and 2 as worse than 3. An omission
leaves the user with the paper prescription, while a
plausible wrong card may be turned into a reminder. The OCR
gate is the response to 4: a poor photograph is rejected
with a request to retake it, not silently processed. We
study the cost of this ordering: a backstop and a gate that
suppress wrong outputs can also suppress right ones, and the
evaluation asks how much.

\textbf{What the backstop assumes.}

\begin{itemize}
\tightlist
\item
  The OCR text contains the medication name as printed or
  written. A drug the OCR never reads cannot be proposed,
  and a drug misread as another plausible word can pass.
\item
  Each medication sits on one line, and its details are near
  its name. This is the assumption that numeric grounding
  breaks on multi-line prescriptions (Section 5.4).
\item
  The name can be located in the OCR text as written. Our
  first backstop required an exact substring, which rejects
  correct names that the OCR spells slightly differently
  (Section 5.7).
\item
  There is no drug-name lexicon. The backstop checks that a
  name is \emph{supported by the text}, not that it is a
  real medication.
\end{itemize}

\textbf{Out of scope.} We do not evaluate the downstream
effect of a reminder, the user interface, or any clinical
outcome. The pipeline is a convenience that proposes, and
every card requires a person's action before anything is
created.

\hypertarget{experimental-design}{%
\section{4 Experimental design}\label{experimental-design}}

Our aim is to find where a small on-device pipeline loses
medications, and which commonly proposed changes recover
them. We therefore hold the input fixed and vary one
pipeline component at a time. This section describes the
test documents, how they were captured, the replay method
that makes the comparison controlled, the configurations,
the metrics, and the statistics. It also states which
comparisons were specified in advance and which were
designed after we had seen the data.

\hypertarget{test-documents}{%
\subsection{4.1 Test documents}\label{test-documents}}

The test set contains 15 synthetic prescriptions, D01 to
D15, for fictional patients. No real patient data were used.
Each prescription lists one to three medications, 24 in
total, with a dose, a frequency code (for example 1-0-1) and
a duration. Appendix A lists the ground truth. Two groups of
documents differ in how the text was produced.

\begin{itemize}
\tightlist
\item
  \textbf{Source A, typed print (D01 to D06).} Printed
  sheets, six documents, 9 medications. Quality tier
  \emph{clear}.
\item
  \textbf{Source C, handwriting (D07 to D15).} The author
  copied the text by hand onto a printed header template
  that carries the clinic name and the doctor's name. Nine
  documents, 15 medications. D07 to D12 are tier
  \emph{average} (everyday handwriting) and D13 to D15 are
  tier \emph{poor} (written fast, letters run together).
\end{itemize}

All handwriting comes from one writer. Source and tier are
fully confounded: every typed sheet is \emph{clear} and
every handwritten sheet is \emph{average} or \emph{poor}, so
a comparison ``by source'' and one ``by tier'' are the same
comparison. We cannot separate the effect of handwriting
from the effect of quality tier, and we do not try to. A
third source, AI-rendered images, was prepared but not
captured, and appears in no result.

Several documents contain deliberate hazards for the
extraction step: clinic and doctor names that look like drug
names, form words (``Tab.'', ``Cap.''), a phone number
beside the vitals, and combination drugs.

\hypertarget{capture-protocol}{%
\subsection{4.2 Capture protocol}\label{capture-protocol}}

Every capture went through the application's own camera
flow, on one phone (Samsung Galaxy F34), in one session, on
a plain surface. We took three captures per document,
varying distance and lighting slightly, and seven for D02.
No sheet carries a document number, tier, or label in the
frame. Three captures of D01 were taken on a cluttered desk.
They are kept but excluded from all results. This leaves
\textbf{49 clean captures} of 15 documents:

\begin{longtable}[]{@{}lccc@{}}
\toprule\noalign{}
Subset & Documents & Captures & Medication slots \\
\midrule\noalign{}
\endhead
\bottomrule\noalign{}
\endlastfoot
A, typed, clear & 6 & 22 & 31 \\
C, handwritten, average & 6 & 18 & 30 \\
C, handwritten, poor & 3 & 9 & 15 \\
\textbf{All} & \textbf{15} & \textbf{49} & \textbf{76} \\
\end{longtable}

A medication \emph{slot} is one ground-truth medication on
one capture, so a prescription with two medications
photographed three times contributes six slots. D02
contributes 7 of the 22 typed captures. We handle this by
resampling at the document level (Section 4.6).

Four live D02 captures were suppressed by the application's
per-category auto-mute. We exclude them from the
live-session results (Section 5.2) and include them in every
replay.

\hypertarget{pipeline-under-test}{%
\subsection{4.3 Pipeline under
test}\label{pipeline-under-test}}

Two pipelines are described in this paper. Sections 5.1 to
5.7 study the pipeline as it stood on 29 October 2026,
described below. On 3 November 2026 it was replaced, and
Section 5.8 reports the replacement. Where the two differ,
Table 11 gives the stages side by side.

The pipeline reads a photograph and proposes medications as
SUGGEST cards, which a user must convert into a reminder or
dismiss. Nothing is executed automatically. In order:

\begin{enumerate}
\def\labelenumi{\arabic{enumi}.}
\tightlist
\item
  \textbf{OCR.} PaddleOCR PP-OCRv5 (an earlier configuration
  uses Tesseract) returns text and a document-level
  confidence.
\item
  \textbf{OCR gate.} A capture whose confidence is below
  \ensuremath{\tau} = 0.87 is rejected and the user is asked
  to retake the photo. The threshold was calibrated on
  earlier pilot captures of the same documents (Section
  5.5).
\item
  \textbf{Extraction.} Gemma 3 270M, quantised, runs on the
  device under a JSON grammar and a repetition penalty, and
  returns a list of medications with dose, frequency and
  duration.
\item
  \textbf{Fabrication backstop.} Deterministic rules drop
  outputs that are not grounded in the OCR text: a stoplist
  of non-medication names, a check that the name appears on
  a line with prescription evidence, removal of few-shot
  placeholder names, and repair of truncated JSON. A
  numeric-grounding step then replaces dose-related fields
  with values found near the name in the OCR text.
\item
  \textbf{Cards.} Surviving outputs become SUGGEST cards.
\end{enumerate}

All generation runs on the phone with the application's own
code. Scoring and statistics are offline.

\hypertarget{frozen-replay-method}{%
\subsection{4.4 Frozen-replay
method}\label{frozen-replay-method}}

Comparing configurations on separate live captures confounds
the pipeline with the camera, the lighting, and the
operator. We avoid this by freezing the input. For every
capture we store the JPEG and the OCR text the application
produced at capture time. A debug-only receiver in the
application then feeds the same frozen text, or the same
JPEG, through any configuration and records the raw model
output and the final cards. OCR noise is therefore identical
across the configurations that share an OCR engine.

Three preconditions justify reading replay results as ``what
the production pipeline would have done on these photos''.
We verified each:

\begin{longtable}[]{@{}
  >{\raggedright\arraybackslash}p{(\columnwidth - 2\tabcolsep) * \real{0.5000}}
  >{\raggedright\arraybackslash}p{(\columnwidth - 2\tabcolsep) * \real{0.5000}}@{}}
\toprule\noalign{}
\begin{minipage}[b]{\linewidth}\raggedright
Check
\end{minipage} & \begin{minipage}[b]{\linewidth}\raggedright
Result
\end{minipage} \\
\midrule\noalign{}
\endhead
\bottomrule\noalign{}
\endlastfoot
Determinism: the production configuration replayed three
times over every capture & 52 of 52 byte-identical raw
outputs \\
Live equals replay: live extractions against replays of the
same capture & 32 of 32 identical (extractions that can be
unambiguously attributed to a capture) \\
OCR reproducibility: PaddleOCR re-run on each frozen JPEG
against the capture-time text & 52 of 52 identical \\
\end{longtable}

Because generation is deterministic, one replay per
configuration carries no sampling variance. All remaining
uncertainty is from the choice of documents (Section 4.6).

The replays of Sections 5.1 to 5.7 use four builds. Within
any table, every row comes from one build. Production output
was byte-identical across the builds on all 52 captures.

The preconditions were verified for the 29 October pipeline.
We do not report them for the replacement pipeline of
Section 5.8, whose results are one run per cell. The
replacement pipeline's model output was generated on the
phone. Its backstop was re-run on the logged model output in
a desktop harness that executes the application's own
backstop code; with the previous backstop code, that harness
reproduces the logged proposals exactly.

\hypertarget{configurations}{%
\subsection{4.5 Configurations}\label{configurations}}

Configurations C1 to C6 and Q5 were specified before the
replay. Each differs from production (C5) in one component,
except where stated.

\begin{longtable}[]{@{}
  >{\raggedright\arraybackslash}p{(\columnwidth - 4\tabcolsep) * \real{0.0789}}
  >{\raggedright\arraybackslash}p{(\columnwidth - 4\tabcolsep) * \real{0.5395}}
  >{\raggedright\arraybackslash}p{(\columnwidth - 4\tabcolsep) * \real{0.3816}}@{}}
\toprule\noalign{}
\begin{minipage}[b]{\linewidth}\raggedright
ID
\end{minipage} & \begin{minipage}[b]{\linewidth}\raggedright
Definition
\end{minipage} & \begin{minipage}[b]{\linewidth}\raggedright
Question it answers
\end{minipage} \\
\midrule\noalign{}
\endhead
\bottomrule\noalign{}
\endlastfoot
C1 & Tesseract OCR, full backstop, Tesseract's own gate
(\ensuremath{\tau} = 0.55) & Does a different OCR engine
matter on identical photos? \\
C2 & PaddleOCR, no backstop, no gate & What does the model
emit unfiltered? \\
C3 & C2 plus the stoplist only & How much does the simplest
filter remove? \\
C4 & C5 with the repetition penalty off & Does the penalty
cost recall? \\
\textbf{C5} & \textbf{Production: PaddleOCR, gate
\ensuremath{\tau} = 0.87, Gemma 3 270M, full backstop} &
\textbf{Reference} \\
C6 & C5 with line-scoped extraction: the model sees only
lines with prescription evidence, plus or minus one line &
Does a shorter, cleaner prompt help? \\
Q5 & C5 with Qwen3-0.6B (Q4\_K\_M) as the extractor & Does a
larger model help? \\
\end{longtable}

Q5 changes the model and the chat template together: Qwen
receives its own template, with the same instructions and
examples in a system turn and \texttt{/no\_think}. The
grammar, backstop and repetition penalty are the same. We
cannot attribute the difference to the model alone (Section
7).

Three further configurations were designed \textbf{after} we
examined the captures and are reported separately as
exploratory (Section 5.7):

\begin{longtable}[]{@{}
  >{\raggedright\arraybackslash}p{(\columnwidth - 2\tabcolsep) * \real{0.1061}}
  >{\raggedright\arraybackslash}p{(\columnwidth - 2\tabcolsep) * \real{0.8939}}@{}}
\toprule\noalign{}
\begin{minipage}[b]{\linewidth}\raggedright
ID
\end{minipage} & \begin{minipage}[b]{\linewidth}\raggedright
Definition
\end{minipage} \\
\midrule\noalign{}
\endhead
\bottomrule\noalign{}
\endlastfoot
C7 & C5 with rotated-box OCR crops, which crop each detected
line as a rotated rectangle rather than an axis-aligned
one \\
C7-T & C7 with tolerant name grounding: a leading
``Tab''/``Cap'' is stripped and a name may match the nearest
OCR word within a length-scaled edit distance (exact for
four characters or fewer, one edit for five to seven, two
for eight or more) \\
Q7-T & Q5 on the rotated-crop OCR text, with tolerant
grounding (handwritten captures only) \\
\end{longtable}

All configurations except C2, which has no gate, are gated.

One further pipeline was adopted after the analyses above
and is evaluated in Section 5.8. We call it \textbf{V2}. It
is not a one-component change from C5: it combines
rotated-box crops, a re-read of each prescription line by a
vision-language model, extraction by that model, the
repetition penalty switched off with a duplicate filter, and
numeric grounding restricted to a drug's own line (Table
11). The model is Qwen2-VL-2B-Instruct {[}Wang et
al.~2024{]}, quantised to Q4\_K\_M, released under the
Apache-2.0 licence. For the ablation of Section 5.8 we add
its components one at a time to the 27 handwritten captures.

\hypertarget{metrics-and-statistics}{%
\subsection{4.6 Metrics and
statistics}\label{metrics-and-statistics}}

\textbf{Metrics.} We score the final SUGGEST cards of each
capture against the ground truth of its document:

\begin{itemize}
\tightlist
\item
  \textbf{Recall:} matched ground-truth slots divided by all
  slots, pooled over captures.
\item
  \textbf{Precision:} matched cards divided by all cards
  emitted.
\item
  \textbf{F1:} harmonic mean of the two.
\item
  \textbf{Character error rate (CER):} the edit distance
  between the text a pipeline produced for a handwritten
  line and our own transcription of that line, divided by
  the length of the transcription. We report it for the 18
  prescription lines, the 16 cursive lines and all 74
  handwritten reference lines. \textbf{Invented words} are
  the share of output words that match no word in the
  transcription.
\item
  \textbf{Unsafe output:} a card whose medication name
  matches no ground-truth medication of its document, for
  example a clinic name, a signature, an advice word, or a
  garbled name.
\item
  \textbf{Dosing error:} a card matched to a ground-truth
  medication whose frequency (times per day) or duration
  (days) differs from the ground truth. D11's Metformin
  frequency is excluded because the ground truth is
  self-contradictory (``0-0-1 (BD)'').
\end{itemize}

A card matches a ground-truth medication when a fuzzy string
match between the two names exceeds a fixed similarity
threshold, which tolerates OCR-level spelling noise. The
match is against the document's own ground-truth list. We
use no external drug lexicon or catalogue. A capture
rejected by the gate emits no cards: its slots count as
missed, and it adds nothing to precision.

\textbf{Failure attribution.} We assign each missed slot to
the first stage that lost it, in this order: rejected by the
gate; name absent from the OCR text; dropped by the backstop
although correct; model emitted a form word; model emitted
header, clinic or advice text; model emitted nothing. The
order makes the categories exclusive.

\textbf{Intervals.} The medication slots within a document,
and the captures of one document, are not independent. We
therefore report 95\% \textbf{document-level cluster
bootstrap} intervals: resample the 15 documents with
replacement, recompute the pooled metric, 10,000 times (seed
20261029). Differences between configurations are
\textbf{paired}: both configurations are scored on the same
resampled documents, which removes the capture-to-capture
variance they share. Wilson intervals, which treat captures
as independent, would be narrower and are not used. With 15
documents the intervals are wide, and a bootstrap over so
few clusters is itself approximate. The same paired
bootstrap is used for differences in character error rate.

\textbf{What the intervals do and do not cover.} They
quantify uncertainty from the choice of documents. They do
not cover the choice of writer, device, or document type,
and they are not out-of-sample estimates, because several
configurations were designed after seeing these captures
(Section 4.7).

\hypertarget{planned-and-post-hoc-analyses}{%
\subsection{4.7 Planned and post-hoc
analyses}\label{planned-and-post-hoc-analyses}}

We separate the analyses by when they were designed.

\begin{itemize}
\item
  \textbf{Specified before the replay:} C1 to C5, Q5, the
  failure taxonomy, the document-level bootstrap, and the
  gate analysis. Even here, the line-scoping rule (C6) and
  the gate threshold were developed on the earlier pilot
  captures of the same documents, so these are in-sample.
\item
  \textbf{Designed after inspecting the replay results:} C7,
  C7-T, Q7-T, and the Rx-presence gate. Each was motivated
  by a failure we saw in these captures, so each evaluation
  on them is in-sample, and the tolerant-grounding gains are
  on exactly the captures that motivated it. We report them
  as a mechanism study. None had changed a production
  default when the 29 October analyses were run; Section 5.8
  reports their adoption.
\item
  \textbf{Adopted after the analyses above:} V2 and its
  ablation (Section 5.8). Its components were chosen after
  inspecting these captures, and it became the production
  pipeline on the strength of this evidence and of earlier
  model comparisons on the phone. It is in-sample
  throughout, and we make no held-out claim for it.
\end{itemize}

No held-out document set exists. Fresh handwriting from a
second writer could not be obtained, so we make \textbf{no
held-out claim} for any number in this paper.

\hypertarget{ethics-and-data}{%
\subsection{4.8 Ethics and data}\label{ethics-and-data}}

All prescriptions are synthetic. The patient, doctor and
clinic names, registration numbers and phone numbers on the
sheets are invented for this study, and any resemblance to
real persons or clinics is unintended. No patient data and
no third-party handwriting are included. The code developed
for this study is for research use only, and no decision has
been made on releasing it. The capture photographs (about
820 MB) are kept by the author and are not distributed. The
test documents and their ground truth are given in Appendix
A. The full prompt, grammar and backstop rule lists are not
reproduced here.

\hypertarget{results}{%
\section{5 Results}\label{results}}

\hypertarget{preconditions}{%
\subsection{5.1 Preconditions}\label{preconditions}}

For the 29 October pipeline, the three preconditions of
Section 4.4 all held: the production configuration was
byte-identical across three repeats on 52 of 52 captures,
live output equalled replay output on 32 of 32 attributable
extractions, and re-running OCR on the frozen JPEGs
reproduced the capture-time text on 52 of 52. Results in
Sections 5.2 to 5.7 can therefore be read as what that
pipeline would have done on these photographs. Because
generation is deterministic, one replay per configuration
suffices. Section 5.8 reports a different pipeline, for
which Section 4.4 states what was and was not rechecked.

\hypertarget{production-behaviour-and-the-configuration-comparison}{%
\subsection{5.2 Production behaviour and the configuration
comparison}\label{production-behaviour-and-the-configuration-comparison}}

\textbf{Live session.} Of the 52 live captures, 32 produced
an extraction, 13 were rejected by the OCR gate, 4 were
suppressed by the auto-mute (D02), and 3 were excluded for
background clutter. Scored on the 45 clean, non-muted
captures:

\begin{longtable}[]{@{}
  >{\raggedright\arraybackslash}p{(\columnwidth - 12\tabcolsep) * \real{0.2917}}
  >{\centering\arraybackslash}p{(\columnwidth - 12\tabcolsep) * \real{0.0833}}
  >{\centering\arraybackslash}p{(\columnwidth - 12\tabcolsep) * \real{0.1389}}
  >{\centering\arraybackslash}p{(\columnwidth - 12\tabcolsep) * \real{0.1667}}
  >{\centering\arraybackslash}p{(\columnwidth - 12\tabcolsep) * \real{0.1667}}
  >{\centering\arraybackslash}p{(\columnwidth - 12\tabcolsep) * \real{0.0694}}
  >{\centering\arraybackslash}p{(\columnwidth - 12\tabcolsep) * \real{0.0833}}@{}}
\toprule\noalign{}
\begin{minipage}[b]{\linewidth}\raggedright
Subset
\end{minipage} & \begin{minipage}[b]{\linewidth}\centering
Captures
\end{minipage} & \begin{minipage}[b]{\linewidth}\centering
Gate pass
\end{minipage} & \begin{minipage}[b]{\linewidth}\centering
Recall
\end{minipage} & \begin{minipage}[b]{\linewidth}\centering
Precision
\end{minipage} & \begin{minipage}[b]{\linewidth}\centering
Unsafe
\end{minipage} & \begin{minipage}[b]{\linewidth}\centering
Dosing errors
\end{minipage} \\
\midrule\noalign{}
\endhead
\bottomrule\noalign{}
\endlastfoot
All & 45 & 71\% (49--91\%) & 6/72 = 8\% (1--17\%) & 67\%
(33--100\%) & 1 & 0 \\
A, typed, clear & 18 & 100\% & 6/27 = 22\% (7--41\%) & 75\%
(54--100\%) & 0 & 0 \\
C, handwritten, average & 18 & 44\% (11--78\%) & 0/30 = 0\%
& 0\% & 1 & 0 \\
C, handwritten, poor & 9 & 67\% & 0/15 = 0\% & none proposed
& 0 & 0 \\
\end{longtable}

Production proposes the right medication for about one in
twelve. The single unsafe card was ``TAB ACEC2OFENAC'' on
D08, a garbled name with a wrong frequency and a 60-day
duration taken from the neighbouring Shelcal line.

\textbf{Configurations on the frozen set.} Table 1 compares
the planned configurations on the 49 clean captures, all
gated except C2. Figure 2 plots the paired differences.

\emph{Table 1. Configurations on identical frozen inputs (49
captures, 76 slots). Recall, precision and F1 with 95\%
document-level intervals; the last column is the paired
difference in recall against C5.}

\begin{longtable}[]{@{}
  >{\raggedright\arraybackslash}p{(\columnwidth - 12\tabcolsep) * \real{0.2500}}
  >{\raggedright\arraybackslash}p{(\columnwidth - 12\tabcolsep) * \real{0.1667}}
  >{\raggedright\arraybackslash}p{(\columnwidth - 12\tabcolsep) * \real{0.1429}}
  >{\raggedright\arraybackslash}p{(\columnwidth - 12\tabcolsep) * \real{0.1429}}
  >{\centering\arraybackslash}p{(\columnwidth - 12\tabcolsep) * \real{0.0595}}
  >{\centering\arraybackslash}p{(\columnwidth - 12\tabcolsep) * \real{0.0595}}
  >{\raggedright\arraybackslash}p{(\columnwidth - 12\tabcolsep) * \real{0.1786}}@{}}
\toprule\noalign{}
\begin{minipage}[b]{\linewidth}\raggedright
Config
\end{minipage} & \begin{minipage}[b]{\linewidth}\raggedright
Recall
\end{minipage} & \begin{minipage}[b]{\linewidth}\raggedright
Precision
\end{minipage} & \begin{minipage}[b]{\linewidth}\raggedright
F1
\end{minipage} & \begin{minipage}[b]{\linewidth}\centering
Unsafe
\end{minipage} & \begin{minipage}[b]{\linewidth}\centering
Dosing errors
\end{minipage} & \begin{minipage}[b]{\linewidth}\raggedright
Paired \ensuremath{\Delta} recall vs C5
\end{minipage} \\
\midrule\noalign{}
\endhead
\bottomrule\noalign{}
\endlastfoot
C1 Tesseract + backstop (\ensuremath{\tau} 0.55) & 2/76 =
3\% (0--6\%) & 100\% & 0.05 (0.02--0.12) & 0 & 0 &
\ensuremath{-}5\% (\ensuremath{-}14 to +2) \\
C2 Paddle, no backstop, no gate & 6/76 = 8\% (1--16\%) &
11\% (2--22\%) & 0.09 & \textbf{45} & 2 & 0 \\
C3 Paddle + stoplist only & 6/76 = 8\% (1--16\%) & 18\%
(4--33\%) & 0.11 & 25 & 2 & 0 \\
C4 repetition penalty off & 13/76 = 17\% (6--29\%) & 81\%
(60--100\%) & 0.28 (0.12--0.43) & 1 & 1 & \textbf{+9\% (+3
to +16)} \\
\textbf{C5 production} & 6/76 = 8\% (1--16\%) & 67\%
(33--100\%) & 0.14 (0.03--0.26) & 1 & 0 & --- \\
C6 line-scoped extraction & 15/76 = 20\% (6--37\%) & 94\%
(80--100\%) & 0.33 (0.12--0.53) & 1 & 2 & \textbf{+12\% (+2
to +24)} \\
Q5 Qwen3-0.6B (model + template) & 28/76 = 37\% (13--62\%) &
90\% (70--100\%) & 0.52 (0.22--0.75) & 3 & 5 & \textbf{+29\%
(+9 to +49)} \\
\end{longtable}

Two observations follow directly. First, C2, C3 and C5 have
the same recall: everything the backstop removed was wrong,
so its cost in recall is zero on this set, while unsafe
outputs fall from 45 to 1. Second, three changes raise
recall with paired intervals above zero: turning the
repetition penalty off, scoping the prompt to prescription
lines, and replacing the model. The Tesseract configuration
is not better than production; the paired difference is
negative and its interval includes zero.

\begin{figure}
\centering
\includegraphics[width=1\textwidth,height=\textheight]{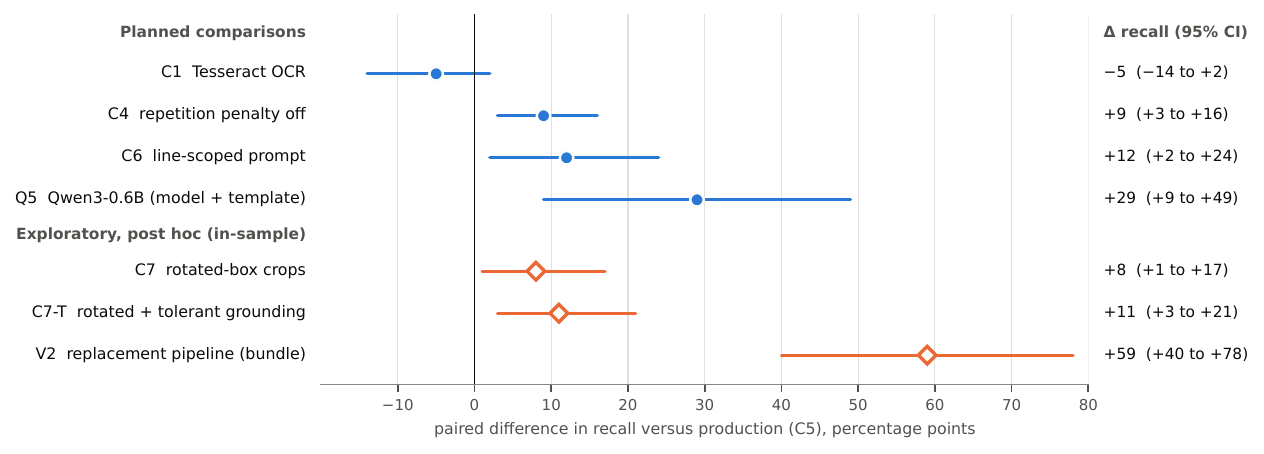}
\caption{Paired difference in recall versus production (C5),
with 95\% document-level bootstrap intervals. Filled circles
are the planned comparisons. Open diamonds are exploratory
comparisons designed after inspecting the captures (Section
5.7); they are in-sample.}
\end{figure}

The live recall for typed sheets (22\%) and the replay
recall (19\%) differ because the live figure excludes the
four muted D02 captures.

\emph{Table 2. Gated recall by source and tier.}

\begin{longtable}[]{@{}
  >{\raggedright\arraybackslash}p{(\columnwidth - 6\tabcolsep) * \real{0.1429}}
  >{\raggedright\arraybackslash}p{(\columnwidth - 6\tabcolsep) * \real{0.3469}}
  >{\raggedright\arraybackslash}p{(\columnwidth - 6\tabcolsep) * \real{0.2653}}
  >{\raggedright\arraybackslash}p{(\columnwidth - 6\tabcolsep) * \real{0.2449}}@{}}
\toprule\noalign{}
\begin{minipage}[b]{\linewidth}\raggedright
Config
\end{minipage} & \begin{minipage}[b]{\linewidth}\raggedright
A typed, clear (31 slots)
\end{minipage} & \begin{minipage}[b]{\linewidth}\raggedright
C average (30)
\end{minipage} & \begin{minipage}[b]{\linewidth}\raggedright
C poor (15)
\end{minipage} \\
\midrule\noalign{}
\endhead
\bottomrule\noalign{}
\endlastfoot
C1 & 6\% (0--13\%) & 0\% & 0\% \\
C4 & 39\% (26--50\%) & 3\% (0--10\%) & 0\% \\
C5 & 19\% (7--36\%) & 0\% & 0\% \\
C6 & 45\% (21--75\%) & 3\% (0--12\%) & 0\% \\
Q5 & \textbf{90\%} (77--100\%) & 0\% & 0\% \\
\end{longtable}

On handwriting no configuration recovers more than one of 45
slots. The differences in Table 1 are therefore differences
on typed text. Because source and tier are confounded
(Section 4.1), Table 2 cannot say whether handwriting or
quality tier explains the gap between its columns.

\hypertarget{where-recall-is-lost}{%
\subsection{5.3 Where recall is
lost}\label{where-recall-is-lost}}

We assign each missed slot to the first stage that lost it
(Section 4.6).

\emph{Table 3. First stage that lost each missed slot.
``Gate'' is rejected by the OCR gate; ``Not in OCR'' is name
absent from the OCR text; ``Backstop'' is a correct name
dropped by the backstop; ``Form word'' and ``Header text''
are model outputs that were a dosage-form word or header,
clinic or advice text; ``Nothing'' is no model output.}

\begin{longtable}[]{@{}
  >{\raggedright\arraybackslash}p{(\columnwidth - 14\tabcolsep) * \real{0.1250}}
  >{\centering\arraybackslash}p{(\columnwidth - 14\tabcolsep) * \real{0.1250}}
  >{\centering\arraybackslash}p{(\columnwidth - 14\tabcolsep) * \real{0.1250}}
  >{\centering\arraybackslash}p{(\columnwidth - 14\tabcolsep) * \real{0.1250}}
  >{\centering\arraybackslash}p{(\columnwidth - 14\tabcolsep) * \real{0.1250}}
  >{\centering\arraybackslash}p{(\columnwidth - 14\tabcolsep) * \real{0.1250}}
  >{\centering\arraybackslash}p{(\columnwidth - 14\tabcolsep) * \real{0.1250}}
  >{\centering\arraybackslash}p{(\columnwidth - 14\tabcolsep) * \real{0.1250}}@{}}
\toprule\noalign{}
\begin{minipage}[b]{\linewidth}\raggedright
Config
\end{minipage} & \begin{minipage}[b]{\linewidth}\centering
Missed
\end{minipage} & \begin{minipage}[b]{\linewidth}\centering
Gate
\end{minipage} & \begin{minipage}[b]{\linewidth}\centering
Not in OCR
\end{minipage} & \begin{minipage}[b]{\linewidth}\centering
Backstop
\end{minipage} & \begin{minipage}[b]{\linewidth}\centering
Form word
\end{minipage} & \begin{minipage}[b]{\linewidth}\centering
Header text
\end{minipage} & \begin{minipage}[b]{\linewidth}\centering
Nothing
\end{minipage} \\
\midrule\noalign{}
\endhead
\bottomrule\noalign{}
\endlastfoot
C5 & 70 & 30 & 16 & \textbf{0} & 6 & 18 & 0 \\
C4 & 63 & 30 & 16 & 0 & 0 & 17 & 0 \\
C6 & 61 & 30 & 15 & 0 & 10 & 6 & 0 \\
Q5 & 48 & 30 & 15 & 0 & 0 & 2 & 1 \\
C1 & 74 & 44 & 0 & 0 & 2 & 28 & 0 \\
\end{longtable}

The losses differ sharply by input type, for the production
configuration:

\begin{itemize}
\tightlist
\item
  \textbf{Handwriting (45 slots).} 30 slots are lost at the
  gate and 13 because the Rx line never appears in the OCR
  text (for example D07, D10, D13, D14). Only 2 slots reach
  the model in readable form. OCR and the gate, not the
  model or the backstop, set the ceiling.
\item
  \textbf{Typed sheets (25 misses).} 22 are model-side: 16
  are header, clinic or advice text taken as a drug, and 6
  are form words. Three are OCR-side (D03 capture 1, D06
  capture 3).
\end{itemize}

The backstop never dropped a correct medication in any
configuration.

\hypertarget{interventions}{%
\subsection{5.4 Interventions}\label{interventions}}

\textbf{Line-scoped extraction (C6).} Sending the model only
the lines carrying prescription evidence, plus or minus one
line, raises recall from 8\% to 20\% (paired +12\%, +2 to
+24) and precision from 67\% to 94\%. It also removes the
D08 unsafe card. It is not uniformly better. On D02, with
the clinic header removed, the model emitted the form word
``Cap.'' on all seven captures, the backstop correctly
dropped it, and nothing was proposed (C5 obtained
Pantoprazole once). C6 also introduced one new unsafe output
on D03 (``Tab.Aiti)'') and three dosing errors through
numeric grounding.

\textbf{Larger model (Q5).} Qwen3-0.6B raises recall to 37\%
(paired +29\%, +9 to +49) and to 90\% on typed sheets. It
does not help on handwriting (0 of 45). The cost is latency
and memory:

\emph{Table 4. On-device cost, Samsung Galaxy F34.}

\begin{longtable}[]{@{}
  >{\raggedright\arraybackslash}p{(\columnwidth - 4\tabcolsep) * \real{0.3247}}
  >{\raggedright\arraybackslash}p{(\columnwidth - 4\tabcolsep) * \real{0.2727}}
  >{\raggedright\arraybackslash}p{(\columnwidth - 4\tabcolsep) * \real{0.4026}}@{}}
\toprule\noalign{}
\begin{minipage}[b]{\linewidth}\raggedright
\end{minipage} & \begin{minipage}[b]{\linewidth}\raggedright
Gemma 3 270M (C5)
\end{minipage} & \begin{minipage}[b]{\linewidth}\raggedright
Qwen3-0.6B Q4\_K\_M (Q5)
\end{minipage} \\
\midrule\noalign{}
\endhead
\bottomrule\noalign{}
\endlastfoot
Model file & 292 MB & 484 MB \\
Median generation time & 12.2 s (p90 14.6 s) & 41.7 s (p90
47.3 s, max 59.6 s) \\
Peak process RSS, median / max & 545 / 626 MB & 666 / 669
MB \\
Unsafe outputs & 1 & 3 (``Advicet'', ``Kemag'',
``Aduice'') \\
Dosing errors & 0 & 5 \\
\end{longtable}

Q5 changes the model and the chat template together, so the
gain cannot be attributed to the model alone (Section 4.5).

\textbf{Repetition penalty (C4).} With the penalty off,
recall rises from 8\% to 17\% (paired +9\%, +3 to +16) and
typed-sheet recall from 19\% to 39\%. The regurgitated
placeholder ``ZolvaX'' returns in 7 raw outputs, but the
backstop's placeholder list removes every one, so unsafe
outputs do not increase (1 versus 1). The penalty was set at
1.3 and never swept. A plausible mechanism is that the
penalty is primed with the whole prompt, which contains the
drug names, and so discourages emitting them. We did not
test this mechanism. The penalty does make generation
slightly faster (median 12.2 s against 14.1 s).

\textbf{Numeric grounding.} The grounding step is not purely
corrective:

\emph{Table 5. Frequency and duration fields changed by
numeric grounding.}

\begin{longtable}[]{@{}lcc@{}}
\toprule\noalign{}
Config & Fields fixed & Fields broken \\
\midrule\noalign{}
\endhead
\bottomrule\noalign{}
\endlastfoot
C5 & 2 & 0 \\
C6 & 10 & 1 \\
Q5 & 5 & 5 \\
\end{longtable}

Across C4, C6 and Q5, grounding introduced 7 of the 8 dosing
errors on correctly named medications. All five Q5 dosing
errors were introduced by grounding, not by the model: D01
Atorvastatin 1\ensuremath{\rightarrow}2 per day because the
±50-character window reaches the Metformin BD line, and D03
Cetirizine 5\ensuremath{\rightarrow}3 days because it
reaches the Azithromycin line. The live D08 60-day error has
the same cause. The window is not scoped to a single
prescription line.

\hypertarget{the-ocr-gate}{%
\subsection{5.5 The OCR gate}\label{the-ocr-gate}}

The document-level gate (\ensuremath{\tau} = 0.87) was
calibrated on 52 earlier PaddleOCR captures of these same
documents, treating the gate as a selective classifier
(Section 4.3). Its structural weakness is that PaddleOCR's
document confidence is a character-weighted mean, so one
garbled drug line moves it little. We evaluated, offline, a
parameter-free alternative that accepts a capture when at
least one prescription line is detected. This check was
designed after inspecting the captures and is exploratory
(Section 4.7). A capture is \emph{adequate} if the drug name
is readable in the OCR text (27 adequate, 22 inadequate).

\emph{Table 6. Gate comparison (offline, 49 captures).}

\begin{longtable}[]{@{}
  >{\raggedright\arraybackslash}p{(\columnwidth - 6\tabcolsep) * \real{0.4861}}
  >{\raggedright\arraybackslash}p{(\columnwidth - 6\tabcolsep) * \real{0.1806}}
  >{\raggedright\arraybackslash}p{(\columnwidth - 6\tabcolsep) * \real{0.1667}}
  >{\raggedright\arraybackslash}p{(\columnwidth - 6\tabcolsep) * \real{0.1667}}@{}}
\toprule\noalign{}
\begin{minipage}[b]{\linewidth}\raggedright
Gate
\end{minipage} & \begin{minipage}[b]{\linewidth}\raggedright
Coverage
\end{minipage} & \begin{minipage}[b]{\linewidth}\raggedright
Selective risk
\end{minipage} & \begin{minipage}[b]{\linewidth}\raggedright
False-reject rate
\end{minipage} \\
\midrule\noalign{}
\endhead
\bottomrule\noalign{}
\endlastfoot
Document \ensuremath{\tau} 0.87 (production) & 0.73 & 0.42 &
0.22 \\
Rx-presence check, no threshold & 0.69 & 0.21 & 0.00 \\
Document \ensuremath{\tau}, leave-one-document-out tuned
(picks 0.91) & 0.41 (0.16--0.65) & 0.00 & 0.26
(0.00--0.62) \\
Rx-presence, leave-one-document-out tuned & 0.61
(0.36--0.83) & 0.20 (0.08--0.39) & 0.11 (0.00--0.38) \\
\end{longtable}

The presence check lowers selective risk by 0.21
(\ensuremath{-}0.42 to \ensuremath{-}0.00) and the
false-reject rate by 0.22 (\ensuremath{-}0.53 to +0.00)
compared with production, but both intervals touch zero.
Eleven of the 15 captures in which no prescription line was
detected still pass \ensuremath{\tau} = 0.87. The
leave-one-out tuned gates trade coverage and risk
differently rather than one dominating. The adequacy label
considers drug names only and ignores dose and frequency
legibility. We did not implement the presence check, and
this analysis is offline only.

\hypertarget{a-retracted-cross-run-result}{%
\subsection{5.6 A retracted cross-run
result}\label{a-retracted-cross-run-result}}

An earlier set of runs, taken in separate sessions with
different capture methods, suggested that a better OCR
engine lowered recall:

\emph{Table 7. Intervals for the earlier runs.}

\begin{longtable}[]{@{}
  >{\raggedright\arraybackslash}p{(\columnwidth - 4\tabcolsep) * \real{0.3095}}
  >{\raggedright\arraybackslash}p{(\columnwidth - 4\tabcolsep) * \real{0.3571}}
  >{\raggedright\arraybackslash}p{(\columnwidth - 4\tabcolsep) * \real{0.3333}}@{}}
\toprule\noalign{}
\begin{minipage}[b]{\linewidth}\raggedright
Run
\end{minipage} & \begin{minipage}[b]{\linewidth}\raggedright
Recall
\end{minipage} & \begin{minipage}[b]{\linewidth}\raggedright
Precision
\end{minipage} \\
\midrule\noalign{}
\endhead
\bottomrule\noalign{}
\endlastfoot
Tesseract (2026-09-24) & 46\% (Wilson 28--65\%) & 69\%
(44--86\%) \\
PaddleOCR, full run & 25\% (Wilson 12--45\%; document
bootstrap 5--50\%) & 60\% (31--83\%) \\
PaddleOCR, final pilot run & 21\% (Wilson 9--40\%; document
bootstrap 7--40\%) & 83\% (Wilson 44--97\%; document
bootstrap 60--100\%) \\
\end{longtable}

The intervals overlap widely, so the difference was never
statistically supported. On identical frozen photographs the
comparison reverses: Tesseract with the full backstop
recalls 3\%, PaddleOCR 8\%, a paired difference of
\ensuremath{-}5\% (\ensuremath{-}14 to +2). Tesseract's own
gate rejects 44 of the 76 slots, and ungated Tesseract still
recalls 3\%. We withdraw the earlier claim. The
capture-method confound is the better explanation, and the
episode is the reason the rest of the study holds the input
fixed.

\hypertarget{exploratory-a-defect-in-the-crop-step-and-what-it-hid}{%
\subsection{5.7 Exploratory: a defect in the crop step, and
what it
hid}\label{exploratory-a-defect-in-the-crop-step-and-what-it-hid}}

\emph{This section reports analyses designed after
inspecting the captures (Section 4.7). They are
mechanism-level, in-sample, and do not support a claim that
the changes generalise.}

\textbf{The defect.} Handwriting recall of 0 of 45 under
every configuration suggested that the OCR stage, not the
model, was the limit. Inspection showed why. Our port of
PaddleOCR cropped each detected text line with an
axis-aligned rectangle padded in proportion to the box area.
On a sloping handwritten line that rectangle is mostly
empty, so the crop became several times taller than the
text. D14's prescription line became a 120-pixel crop of a
50-pixel line plus the ``Rx'' heading, which the recogniser
read as blank. This is a defect in our port, not in
PaddleOCR or the model. Configuration C7 replaces the crop
with PaddleOCR's own method: a rotated rectangle, its score,
unclipping on the rotated box, and a perspective crop.

\textbf{Effect on OCR.}

\emph{Table 8. Drug names recoverable from the OCR text (49
captures).}

\begin{longtable}[]{@{}
  >{\raggedright\arraybackslash}p{(\columnwidth - 6\tabcolsep) * \real{0.1935}}
  >{\centering\arraybackslash}p{(\columnwidth - 6\tabcolsep) * \real{0.1452}}
  >{\raggedright\arraybackslash}p{(\columnwidth - 6\tabcolsep) * \real{0.4194}}
  >{\raggedright\arraybackslash}p{(\columnwidth - 6\tabcolsep) * \real{0.2419}}@{}}
\toprule\noalign{}
\begin{minipage}[b]{\linewidth}\raggedright
Subset
\end{minipage} & \begin{minipage}[b]{\linewidth}\centering
Axis-aligned
\end{minipage} & \begin{minipage}[b]{\linewidth}\raggedright
Rotated
\end{minipage} & \begin{minipage}[b]{\linewidth}\raggedright
Gate pass at \ensuremath{\tau} 0.87, axis
\ensuremath{\rightarrow} rotated
\end{minipage} \\
\midrule\noalign{}
\endhead
\bottomrule\noalign{}
\endlastfoot
A typed & 28/31 & 31/31 & 22 \ensuremath{\rightarrow} 22 of
22 \\
C average & 13/30 & 19/30 & 8 \ensuremath{\rightarrow} 13 of
18 \\
C poor & 1/15 & \textbf{15/15} & 6 \ensuremath{\rightarrow}
5 of 9 \\
C all & 14/45 & \textbf{34/45} (document-bootstrap interval
of the difference 0--77\%) & 14 \ensuremath{\rightarrow} 18
of 27 \\
\end{longtable}

Four captures got worse: D09 captures 1 and 2 lose
Pantoprazole, and D10 capture 2 and D14 capture 2 fall below
the gate. OCR time rises from 1.84 s to 2.06 s per
photograph (median).

\textbf{End-to-end effect.}

\emph{Table 9. Production versus rotated crops (gated, 49
captures).}

\begin{longtable}[]{@{}
  >{\raggedright\arraybackslash}p{(\columnwidth - 6\tabcolsep) * \real{0.2586}}
  >{\raggedright\arraybackslash}p{(\columnwidth - 6\tabcolsep) * \real{0.2414}}
  >{\raggedright\arraybackslash}p{(\columnwidth - 6\tabcolsep) * \real{0.2414}}
  >{\raggedright\arraybackslash}p{(\columnwidth - 6\tabcolsep) * \real{0.2586}}@{}}
\toprule\noalign{}
\begin{minipage}[b]{\linewidth}\raggedright
\end{minipage} & \begin{minipage}[b]{\linewidth}\raggedright
C5
\end{minipage} & \begin{minipage}[b]{\linewidth}\raggedright
C7 rotated
\end{minipage} & \begin{minipage}[b]{\linewidth}\raggedright
C7-T rotated + tolerant
\end{minipage} \\
\midrule\noalign{}
\endhead
\bottomrule\noalign{}
\endlastfoot
Recall & 6/76 = 8\% (1--16\%) & 12/76 = 16\% (5--28\%) &
14/76 = 18\% (8--31\%) \\
Paired \ensuremath{\Delta} vs C5 & --- & +8\% (+1 to +17) &
\textbf{+11\% (+3 to +21)} \\
Precision & 67\% & 86\% (64--100\%) & 88\% \\
Unsafe & 1 & 0 & 0 \\
Typed recall & 19\% & 35\% (20--50\%) & unchanged from C7 \\
Handwritten recall & 0/45 & 1/45 & 3/45 \\
\end{longtable}

\textbf{The bottleneck moves.} Under C7, the 45 handwritten
slots end as follows (Figure 3, panel b):

\begin{longtable}[]{@{}
  >{\raggedright\arraybackslash}p{(\columnwidth - 2\tabcolsep) * \real{0.9077}}
  >{\centering\arraybackslash}p{(\columnwidth - 2\tabcolsep) * \real{0.0923}}@{}}
\toprule\noalign{}
\begin{minipage}[b]{\linewidth}\raggedright
Where the slot ends up
\end{minipage} & \begin{minipage}[b]{\linewidth}\centering
Slots
\end{minipage} \\
\midrule\noalign{}
\endhead
\bottomrule\noalign{}
\endlastfoot
Not in the OCR text & 11 \\
Readable, but the gate rejects the capture (confidence
0.85--0.87) & 17 \\
Readable and passes the gate, but Gemma names something else
(clinic, doctor, ``Korla'', ``ZolvaX'') & 14 \\
Gemma correct, but the backstop drops it & 2 \\
Matched & 1 \\
\end{longtable}

\begin{figure}
\centering
\includegraphics[width=1\textwidth,height=\textheight]{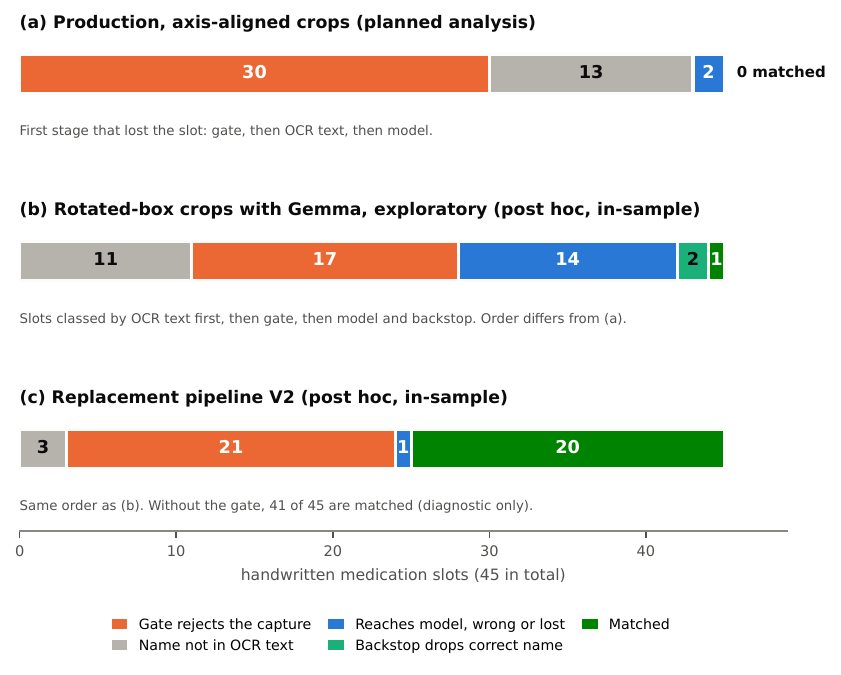}
\caption{Where the 45 handwritten medication slots end up.
(a) Production, each slot assigned to the first stage that
lost it (gate, then OCR text, then model). (b) Exploratory
rotated-box crops (C7), with slots classified by OCR text
first, then gate, then model and backstop. (c) The
replacement pipeline of Section 5.8, classified in the same
order as (b). The order of (a) differs, so its buckets are
not directly comparable with (b) and (c). Counts are slots;
in-sample.}
\end{figure}

The two backstop drops are on D14: Gemma wrote
``Nitrofurantoin'' where the OCR reads ``Nitrofurantonin'',
and the exact-substring check rejected the correctly spelled
name. Repairing OCR therefore exposes three further limits,
in order: a gate threshold calibrated on axis-aligned OCR,
the extractor's choice of text, and exact-match grounding.

\textbf{Tolerant grounding.} Configuration C7-T strips a
leading ``Tab'' or ``Cap'' and accepts a name within a
length-scaled edit distance of an OCR word (Section 4.5). It
left every typed result unchanged and removed nothing that
exact grounding kept. On handwriting it raised Gemma from 1
to 3 of 45 (paired +4\%, 0 to +17). Every added match is a
case that motivated the change: D14 Nitrofurantoin for
Gemma, and D12 Naproxen plus D14 Nitrofurantoin for Qwen. We
treat these gains as illustrations of the mechanism, not as
evidence of improvement.

\textbf{Qwen on rotated text.} Run on the 27 handwritten
captures:

\emph{Table 10. Handwritten captures, rotated OCR (45
slots).}

\begin{longtable}[]{@{}
  >{\raggedright\arraybackslash}p{(\columnwidth - 4\tabcolsep) * \real{0.4366}}
  >{\raggedright\arraybackslash}p{(\columnwidth - 4\tabcolsep) * \real{0.2113}}
  >{\raggedright\arraybackslash}p{(\columnwidth - 4\tabcolsep) * \real{0.3521}}@{}}
\toprule\noalign{}
\begin{minipage}[b]{\linewidth}\raggedright
\end{minipage} & \begin{minipage}[b]{\linewidth}\raggedright
Gemma 3 270M
\end{minipage} & \begin{minipage}[b]{\linewidth}\raggedright
Qwen3-0.6B
\end{minipage} \\
\midrule\noalign{}
\endhead
\bottomrule\noalign{}
\endlastfoot
Gated recall, exact grounding & 1/45 = 2\% (0--7\%) & 3/45 =
7\% (0--18\%) \\
Paired \ensuremath{\Delta} recall & --- & +4\%
(\ensuremath{-}5 to +17), not significant \\
Gated precision & 100\% (one output) & 33\% (0--90\%) \\
Unsafe outputs & 0 & 6 \\
Raw model output matching the drug, fuzzy, before the
backstop & 4/45 & \textbf{11/45} \\
Gated recall, tolerant grounding & 3/45 & 7/45 = 16\%
(4--36\%), 7 unsafe \\
Median generation time & \textasciitilde12 s & 39.9 s (max
69.3 s) \\
\end{longtable}

Qwen names the drug more often than Gemma, but it copies the
OCR's misspellings or fuses them with the dosage form, and
the exact-match backstop drops 7 of its 11 near-correct
names. Tolerant grounding recovers some of them but also
lets one more garbled output through (``CETI R121 NE'', D10
capture 1). Qwen's signature and advice-line outputs still
pass the backstop, so precision on handwriting stays near
50\% without a drug-name lexicon. On this data, swapping the
model does not fix handwriting: the next limit is grounding.

\hypertarget{replacing-the-model-a-vision-language-extractor}{%
\subsection{5.8 Replacing the model: a vision-language
extractor}\label{replacing-the-model-a-vision-language-extractor}}

\emph{This section reports changes made after the analyses
above and evaluated on the same captures. Several were
designed after inspecting them, and the changes were adopted
together. It is in-sample, and it is a before-and-after
comparison of two pipelines, not a controlled comparison of
one component.}

\textbf{What changed.} On 3 November 2026 we replaced both
Gemma 3 270M and Qwen3-0.6B with Qwen2-VL-2B-Instruct
(Q4\_K\_M), a 2-billion-parameter vision-language model, and
rebuilt the pipeline around it (Table 11). The model has two
roles. It re-reads each line that carries prescription
evidence, plus or minus one line, from the image at 64
pixels height, and it then extracts the medications from the
resulting text.

\emph{Table 11. Production pipeline before and after.}

\begin{longtable}[]{@{}
  >{\raggedright\arraybackslash}p{(\columnwidth - 4\tabcolsep) * \real{0.2239}}
  >{\raggedright\arraybackslash}p{(\columnwidth - 4\tabcolsep) * \real{0.3433}}
  >{\raggedright\arraybackslash}p{(\columnwidth - 4\tabcolsep) * \real{0.4328}}@{}}
\toprule\noalign{}
\begin{minipage}[b]{\linewidth}\raggedright
Stage
\end{minipage} & \begin{minipage}[b]{\linewidth}\raggedright
Before (Section 4.3)
\end{minipage} & \begin{minipage}[b]{\linewidth}\raggedright
After
\end{minipage} \\
\midrule\noalign{}
\endhead
\bottomrule\noalign{}
\endlastfoot
OCR crops & Axis-aligned & Rotated boxes (Section 5.7) \\
OCR gate & \ensuremath{\tau} = 0.87 on document confidence &
Unchanged. We chose not to loosen it. \\
Reading handwriting & OCR text only & Qwen2-VL-2B re-reads
prescription lines from the image \\
Extractor & Gemma 3 270M, repetition penalty on &
Qwen2-VL-2B, penalty off, duplicate filter \\
Backstop & Full & Full; numeric grounding reads each drug's
own line only \\
\end{longtable}

\textbf{Before and after.} Table 12 scores the old
production configuration (C5) against the new pipeline on
the same 49 captures, gated as in production. The model
output was generated on the phone. The per-line
numeric-grounding fix was applied to the logged model output
in a desktop harness that runs the application's backstop
code; with the old code, the harness reproduces the logged
proposals exactly.

\emph{Table 12. Production before and after (gated, 49
captures, 76 slots).}

\begin{longtable}[]{@{}
  >{\raggedright\arraybackslash}p{(\columnwidth - 6\tabcolsep) * \real{0.3151}}
  >{\raggedright\arraybackslash}p{(\columnwidth - 6\tabcolsep) * \real{0.2192}}
  >{\raggedright\arraybackslash}p{(\columnwidth - 6\tabcolsep) * \real{0.2329}}
  >{\raggedright\arraybackslash}p{(\columnwidth - 6\tabcolsep) * \real{0.2329}}@{}}
\toprule\noalign{}
\begin{minipage}[b]{\linewidth}\raggedright
Subset
\end{minipage} & \begin{minipage}[b]{\linewidth}\raggedright
Recall before
\end{minipage} & \begin{minipage}[b]{\linewidth}\raggedright
Recall after
\end{minipage} & \begin{minipage}[b]{\linewidth}\raggedright
Paired \ensuremath{\Delta} (95\% CI)
\end{minipage} \\
\midrule\noalign{}
\endhead
\bottomrule\noalign{}
\endlastfoot
All & 6/76 = 8\% (1--16\%) & \textbf{51/76 = 67\%}
(45--89\%) & \textbf{+59} (+40 to +78) \\
Typed (22 captures, 31 slots) & 6/31 = 19\% (7--36\%) &
\textbf{31/31 = 100\%} & \textbf{+81} (+64 to +93) \\
Handwritten (27 captures, 45 slots) & 0/45 & \textbf{20/45 =
44\%} (23--73\%) & \textbf{+44} (+23 to +73) \\
\end{longtable}

Overall precision rises from 67\% to 96\%, and unsafe
outputs fall from 1 to 0. Dosing errors on matched
medications are 3 of 51 gated matches (6 of 72 ungated).
Before the per-line grounding fix the same run had 12 gated
(17 ungated), so the fix removes the mechanism behind most
dosing errors in Section 5.4. Of the remaining three, one
comes from D09, where the OCR reading order places both
drugs' dosing lines after the second drug, and one from a
D08 duration written without the word ``days''. The model's
raw output names no clinic, hospital or doctor on any of the
49 captures.

\textbf{Where the handwriting gain comes from.} Three
changes arrived together on handwriting: rotated crops (an
OCR fix), the re-read (the model's vision) and the
extraction (its language side). We added them one at a time
on the 27 handwritten captures. The reading step is scored
by character error rate (CER) against our own transcription
of the 74 handwritten reference lines.

\emph{Table 13. Reading accuracy of handwritten text.}

\begin{longtable}[]{@{}
  >{\raggedright\arraybackslash}p{(\columnwidth - 8\tabcolsep) * \real{0.4375}}
  >{\centering\arraybackslash}p{(\columnwidth - 8\tabcolsep) * \real{0.1406}}
  >{\centering\arraybackslash}p{(\columnwidth - 8\tabcolsep) * \real{0.1406}}
  >{\centering\arraybackslash}p{(\columnwidth - 8\tabcolsep) * \real{0.1406}}
  >{\centering\arraybackslash}p{(\columnwidth - 8\tabcolsep) * \real{0.1406}}@{}}
\toprule\noalign{}
\begin{minipage}[b]{\linewidth}\raggedright
Text given to the extractor
\end{minipage} & \begin{minipage}[b]{\linewidth}\centering
Drug names present
\end{minipage} & \begin{minipage}[b]{\linewidth}\centering
CER, prescription lines
\end{minipage} & \begin{minipage}[b]{\linewidth}\centering
CER, all lines
\end{minipage} & \begin{minipage}[b]{\linewidth}\centering
Invented words
\end{minipage} \\
\midrule\noalign{}
\endhead
\bottomrule\noalign{}
\endlastfoot
Axis-aligned OCR & 14/45 & 52.6\% & 31.3\% & 8.4\% \\
Rotated-box OCR & 34/45 & 15.9\% & 12.0\% & 7.8\% \\
Rotated OCR + Qwen2-VL re-read & \textbf{42/45} &
\textbf{10.1\%} & \textbf{9.8\%} & \textbf{4.8\%} \\
\end{longtable}

Rotated crops account for most of the reading gain
(prescription-line CER \ensuremath{-}36.7 points, interval
\ensuremath{-}54.4 to \ensuremath{-}15.5). The re-read adds
a smaller but clear gain on the lines it reads
(\ensuremath{-}5.8 points, \ensuremath{-}7.4 to
\ensuremath{-}3.9), and nearly halves invented words.

\emph{Table 14. Medications proposed correctly on
handwriting (45 slots).}

\begin{longtable}[]{@{}
  >{\raggedright\arraybackslash}p{(\columnwidth - 6\tabcolsep) * \real{0.5500}}
  >{\centering\arraybackslash}p{(\columnwidth - 6\tabcolsep) * \real{0.1500}}
  >{\centering\arraybackslash}p{(\columnwidth - 6\tabcolsep) * \real{0.1500}}
  >{\centering\arraybackslash}p{(\columnwidth - 6\tabcolsep) * \real{0.1500}}@{}}
\toprule\noalign{}
\begin{minipage}[b]{\linewidth}\raggedright
Pipeline
\end{minipage} & \begin{minipage}[b]{\linewidth}\centering
Gated
\end{minipage} & \begin{minipage}[b]{\linewidth}\centering
Ungated
\end{minipage} & \begin{minipage}[b]{\linewidth}\centering
Unsafe (gated / ungated)
\end{minipage} \\
\midrule\noalign{}
\endhead
\bottomrule\noalign{}
\endlastfoot
Gemma, axis-aligned OCR (old production) & 0/45 & 0/45 & 1 /
1 \\
Gemma, rotated OCR & 1/45 & 2/45 & 0 / 2 \\
Qwen2-VL, rotated OCR & 12/45 = 27\% & 24/45 = 53\% & 2 /
5 \\
\textbf{Qwen2-VL, rotated OCR + re-read} & \textbf{20/45 =
44\%} & \textbf{41/45 = 91\%} & \textbf{0 / 0} \\
\end{longtable}

The paired gated gains are +2 for rotated crops with Gemma
(0 to +7), \textbf{+24} for Qwen2-VL extraction in place of
Gemma on the same text (+5 to +51), and \textbf{+18} for the
re-read with the same extractor (+3 to +39). The two
readings answer different questions. Reading accuracy
improved mostly because of the crop fix. Medications found
improved mostly because of the model: with the old
extractor, 34 of 45 names were present in the text and Gemma
extracted 2, naming the clinic or the doctor instead, while
Qwen2-VL extracted 24 from the same imperfect text. The
re-read also cleans the proposals: Qwen2-VL on rotated OCR
alone made five unsafe ungated proposals (four garbled
copies of the right drug, and ``Advice'' read as
``Aduice''), and with the re-read there are none.

\textbf{The gate is now the limit} (Figure 3, panel c). Of
the 25 handwritten slots the new pipeline misses, 21 are
readable but sit on captures the gate rejects, 3 are not in
the text, and 1 is readable and gated in but not extracted.
Without the gate the same pipeline finds 41 of 45
handwritten slots (91\%, 72--100\%) and 72 of 76 overall
(95\%), still with no unsafe output. We kept
\ensuremath{\tau} = 0.87 deliberately, because it guards
against garbled OCR becoming wrong medications; we give the
ungated figures as a diagnostic, not as production
behaviour. This matches the prediction of Section 5.7, that
repairing the crops would move the bottleneck to the gate
and the extractor.

\textbf{Cost.} The gain is paid for in time, memory and
size.

\emph{Table 15. Cost on the test phone.}

\begin{longtable}[]{@{}
  >{\raggedright\arraybackslash}p{(\columnwidth - 4\tabcolsep) * \real{0.3867}}
  >{\raggedright\arraybackslash}p{(\columnwidth - 4\tabcolsep) * \real{0.2267}}
  >{\raggedright\arraybackslash}p{(\columnwidth - 4\tabcolsep) * \real{0.3867}}@{}}
\toprule\noalign{}
\begin{minipage}[b]{\linewidth}\raggedright
\end{minipage} & \begin{minipage}[b]{\linewidth}\raggedright
Before
\end{minipage} & \begin{minipage}[b]{\linewidth}\raggedright
After
\end{minipage} \\
\midrule\noalign{}
\endhead
\bottomrule\noalign{}
\endlastfoot
Time per photograph, median (p90) & \textasciitilde12 s
after OCR & 61 s (87 s): OCR 2.6 s, re-read 42.5 s,
extraction 12.1 s \\
Memory while reading & \textasciitilde545 MB peak &
\textasciitilde1.9 GB while re-reading, \textasciitilde1.2
GB while extracting \\
Application size & 1.18 GB & 2.12 GB (model 986 MB, vision
projector 710 MB) \\
Low-memory kills while in front & none seen & 3 before
memory fixes, 0 in the 49-capture run after \\
\end{longtable}

The application must stay on screen while a photograph is
read. With the screen locked the app is throttled to the
background CPU group, and each line took 55 to 217 s instead
of 8 to 13 s.

\textbf{What this does not show.} It does not show that a
2-billion-parameter vision-language model generalises to
other writers. Every number is in-sample: the same 15
documents, with design choices made after we had seen them.
The comparison bundles six changes (crops, re-read, model,
penalty, duplicate filter, grounding fix), and the ablation
covers handwriting and the first three. Gemma was not run on
the re-read text, so the re-read's effect is measured with
Qwen2-VL only. The model and the prompt template change
together between the Gemma and Qwen2-VL rows, as in Section
4.5. The CER references are our own transcriptions. We do
not report the determinism and live-equals-replay checks of
Section 5.1 for this pipeline, so Table 12 is one run per
cell.

\hypertarget{discussion}{%
\section{6 Discussion}\label{discussion}}

\hypertarget{a-flawed-upstream-stage-can-hide-every-downstream-comparison}{%
\subsection{6.1 A flawed upstream stage can hide every
downstream
comparison}\label{a-flawed-upstream-stage-can-hide-every-downstream-comparison}}

The clearest lesson is about method. Under configurations C1
to C6, handwriting recall was 0 or 1 of 45, so no change to
the sampler, the prompt, or the model could be told apart on
handwriting. The cause was a crop step in our own OCR port
that read sloping lines as blank (Section 5.7). Only the
stage-attributed taxonomy (Section 5.3), which asks where
each slot was lost, pointed away from the model and toward
the input. Comparing mitigations on the final output alone
would have said that nothing helps handwriting. A study that
reports end-to-end recall without attributing losses to
stages cannot distinguish ``this component does not help''
from ``this component never received usable input''.

The replacement pipeline of Section 5.8 makes the same point
from the other side. Once the crops were repaired, a change
of model that earlier did nothing for handwriting
(Qwen3-0.6B, Table 10) became visible, and a stronger model
then added a gain that only an ablation could attribute.
Without Tables 13 and 14 the rise from 8\% to 67\% would
read as ``the bigger model works''. It is two different
things: the crop fix accounts for most of the improvement in
reading accuracy, and the model for most of the improvement
in medications found.

\hypertarget{typed-text-is-a-model-and-prompt-problem}{%
\subsection{6.2 Typed text is a model and prompt
problem}\label{typed-text-is-a-model-and-prompt-problem}}

Once text is readable, the losses on typed sheets are on the
model side: 22 of 25 misses were header, clinic or advice
text, or a form word taken as the drug (Section 5.3). Three
changes helped. Restricting the prompt to prescription lines
raised recall from 8\% to 20\% without a larger model.
Replacing Gemma with Qwen3-0.6B raised typed-sheet recall to
90\% at three times the generation time (41.7 s against 12.2
s). The replacement pipeline of Section 5.8 reached 31 of 31
typed slots, and its raw output named no clinic, hospital or
doctor on any capture. We cannot say how much of that comes
from the larger model, how much from its prompt template and
how much from the re-read, because typed text was not
ablated. The scoped prompt is the cheapest change. None of
these results should be read as a recommendation without a
held-out test.

\hypertarget{safeguards-interact-and-some-become-redundant}{%
\subsection{6.3 Safeguards interact, and some become
redundant}\label{safeguards-interact-and-some-become-redundant}}

The repetition penalty was added to stop the model copying
the prompt's example drug. After the placeholder filter
existed, the penalty was no longer needed for safety and was
costing recall (Section 5.4). Numeric grounding was meant to
repair dose fields and did repair some (10 fields in C6),
but its fixed window crossed into neighbouring lines and
broke others, and accounted for 7 of the 8 dosing errors on
correctly named medications. Each safeguard looked sensible
when added. Only the ablation against the final pipeline
showed one to be net negative and another to be
double-edged. We suggest that a safeguard be re-evaluated
whenever another is added in front of or behind it.

\hypertarget{strict-grounding-trades-recall-for-safety-and-the-price-rises-as-ocr-improves}{%
\subsection{6.4 Strict grounding trades recall for safety,
and the price rises as OCR
improves}\label{strict-grounding-trades-recall-for-safety-and-the-price-rises-as-ocr-improves}}

The backstop removed no correct medication in any
configuration, so its cost in recall was zero on this set.
That held because the model's correct answers matched the
OCR text exactly. Once handwriting became readable,
exact-substring grounding rejected correctly named drugs
that the OCR misspelled (Nitrofurantoin against
``Nitrofurantonin''), and 7 of Qwen's 11 near-correct names
(Section 5.7). Tolerant matching recovered some of them, and
also admitted one more garbled output. The tension is
structural: grounding that tolerates OCR error also
tolerates model error. A drug-name lexicon would help
separate the two, and we did not test one.

\hypertarget{a-document-level-confidence-is-the-wrong-summary-for-a-gate}{%
\subsection{6.5 A document-level confidence is the wrong
summary for a
gate}\label{a-document-level-confidence-is-the-wrong-summary-for-a-gate}}

PaddleOCR's document confidence is a character-weighted
mean, so one garbled drug line barely moves it while the
rest of the page is clean. Eleven of 15 captures with no
prescription line detected still passed the production
threshold. A check on the field that matters, whether a
prescription line was found, did better on both selective
risk and false-reject rate in this sample, though both
intervals touch zero (Section 5.5). The general point is
that a confidence gate should look at the part of the input
that the downstream step depends on, not at an average over
the page.

In the replacement pipeline the gate is the largest
remaining loss: 21 of the 25 handwritten slots still missed
are readable but sit on captures the gate rejects, and
without it the pipeline finds 41 of 45. We kept
\ensuremath{\tau} = 0.87 on purpose, because the threshold
guards against garbled OCR becoming wrong medications. Our
gate analysis (Section 5.5) suggests that a presence check
on prescription lines would do better, but it was not tested
on the new pipeline, and its interval touches zero.

\hypertarget{what-a-deployment-might-do}{%
\subsection{6.6 What a deployment might
do}\label{what-a-deployment-might-do}}

On this data, the 29 October pipeline would often decline a
handwritten prescription and often miss medications on a
typed one. The replacement reads far more, but it takes
about a minute per photograph, needs about 1.9 GB of memory
while re-reading, adds about a gigabyte to the application,
and must stay on screen, because a locked screen slowed each
line from 8 to 13 s to 55 to 217 s. These costs fall on a
user who is waiting for a result. The design choice in
Section 3.3 still makes the failures acceptable, since a
missed medication leaves the user with the original, while a
wrong card does not. The system should say clearly what it
did not read (the silent failure of Section 3.3), and a
deployment should treat the cards as a convenience check
against the paper, never a substitute for it.

\hypertarget{two-levers-on-handwriting}{%
\subsection{6.7 Two levers on
handwriting}\label{two-levers-on-handwriting}}

On handwriting, reading and extracting are different
problems with different remedies. Reading failed because the
crop step handed the recogniser mostly empty rectangles, and
rotated crops fixed that regardless of the model: the
prescription-line character error rate fell from 53\% to
16\% with the same OCR engine. Extracting failed because
Gemma, given text in which 34 of 45 drug names were present,
named the clinic or doctor instead and extracted 2. A larger
model extracted 24 of the same 45 from the same imperfect
text, and a re-read of the lines by that model added a
further gain and nearly halved invented words. A study that
varied only the model on the old crops would have found
nothing (Table 10), and one that varied only the crops would
have found little (Table 14, +2 with Gemma). The gain exists
only when both are changed. We read this as a general
caution for pipelines of this shape: a stage that looks like
the bottleneck may be masked by the one before it.

\hypertarget{future-work}{%
\subsection{6.8 Future work}\label{future-work}}

Each item below follows from a limit in Section 7.

\begin{itemize}
\tightlist
\item
  A held-out evaluation with several writers, several
  devices and real prescriptions under proper ethical
  approval, to test whether the post-hoc changes of Section
  5.7 and the replacement pipeline of Section 5.8 hold.
\item
  A template-swap control to separate each model from its
  chat template, and a run of Gemma on the re-read text to
  separate the re-read from the model.
\item
  Ablating the re-read and the extraction on typed text,
  which Section 5.8 did not do.
\item
  A drug-name lexicon for grounding, and a presence check or
  re-calibrated gate for the new pipeline.
\item
  Reducing the latency and memory cost of the re-read, for
  example by re-reading only the lines the recogniser is
  unsure of.
\item
  Using the public prescription datasets (word-level
  handwriting data and document-level sets) to test the OCR
  stage at scale, which our 15 documents cannot.
\end{itemize}

\hypertarget{threats-to-validity}{%
\section{7 Threats to validity}\label{threats-to-validity}}

The study is a controlled comparison on a small,
single-source test set. We state the limits in the order a
reader would use them to discount a result: what the data
cannot support, what could bias it, and what we could not
check.

\hypertarget{no-held-out-data}{%
\subsection{7.1 No held-out data}\label{no-held-out-data}}

Every number in this paper is in-sample. A second writer's
handwriting could not be obtained, so no document in the
test set was kept aside. This matters most for the analyses
designed after we had seen the captures: rotated-box crops,
tolerant name grounding, and the prescription-line presence
check (Section 4.7). Each was motivated by a failure in
these 49 captures, and the tolerant-grounding gains fall on
exactly those captures. Line-scoped extraction and the gate
threshold were developed on earlier pilot captures of the
same 15 documents. We therefore make no claim that any of
these changes would help on other writers, devices, or
documents, and the 29 October evaluation changed no
production default. A reader should treat the gains in
Section 5.7 as evidence of a mechanism, not of an
improvement.

The replacement pipeline of Section 5.8 has the same limit.
Its components were chosen after we had seen these captures,
and it became the production default. That is a decision to
adopt it, not a validation of it.

\hypertarget{writer-documents-and-device}{%
\subsection{7.2 Writer, documents and
device}\label{writer-documents-and-device}}

The handwriting is one person's, written to copy fictional
prescriptions onto a printed header template. Some header
slips were kept as they were. The typed sheets were printed
in the author's own layout. Real prescriptions vary in
layout, abbreviations, stamps, pre-printed fields, language,
and writing style, and many are written under time pressure
by someone who has never seen the pipeline. The test set has
24 medications, all common, with one dose-code convention
(for example 1-0-1). All captures use one phone camera, one
OCR engine family, and one on-device runtime. Our
\emph{poor} tier is an imitation of fast handwriting by the
author, not an observed sample of it.

\hypertarget{confounds-we-could-not-separate}{%
\subsection{7.3 Confounds we could not
separate}\label{confounds-we-could-not-separate}}

\begin{itemize}
\tightlist
\item
  \textbf{Source and tier.} Every typed sheet is
  \emph{clear} and every handwritten sheet is \emph{average}
  or \emph{poor} (Section 4.1). The study cannot say whether
  handwriting itself or lower image quality explains the gap
  between Table 2's columns.
\item
  \textbf{Model and chat template.} Qwen3-0.6B was run with
  its own chat template (Section 4.5). The +29\% gain of Q5
  belongs to the pair, and the template may account for part
  of it. We did not run the swap that would separate them.
\item
  \textbf{Bundled change.} The replacement pipeline combines
  six changes (Table 11). The ablation of Section 5.8 adds
  three of them one at a time, on handwriting only. The
  repetition penalty, the duplicate filter and the grounding
  fix are not separated, and nothing on typed text is
  ablated.
\item
  \textbf{Model and template again.} Gemma and Qwen2-VL
  differ in prompt template as well as in model, and Gemma
  was never run on the re-read text. The re-read's effect is
  measured with Qwen2-VL only.
\item
  \textbf{The reference for reading accuracy.} Character
  error rates are against the author's own transcription of
  the handwriting, so the author both wrote and transcribed
  the lines.
\item
  \textbf{A defect in our own OCR port.} The 0 of 45
  handwriting baseline measures a port with a flawed crop
  step (Section 5.7), not PaddleOCR in general. Statements
  about handwriting under C1 to C6 are statements about that
  port.
\item
  \textbf{Repetition penalty.} We report that turning the
  penalty off raises recall. Our explanation, that the
  prompt primes the penalty with the drug names, is a
  hypothesis we did not test.
\end{itemize}

\hypertarget{measurement}{%
\subsection{7.4 Measurement}\label{measurement}}

\begin{itemize}
\tightlist
\item
  \textbf{Metric definitions.} Recall is pooled over slots,
  so documents with more medications and more captures weigh
  more; D02 alone supplies 7 of 22 typed captures. The
  document-level bootstrap mitigates but does not remove
  this.
\item
  \textbf{``Unsafe'' is a name-level rule.} An output is
  unsafe when it names no ground-truth medication (Section
  4.6). It does not measure clinical harm, and a correct
  name with a wrong dose is counted separately as a dosing
  error. We make no claim about clinical accuracy or
  readiness.
\item
  \textbf{Adequacy labels} (used for the gate analysis)
  consider drug names only and ignore the legibility of dose
  and frequency.
\item
  \textbf{Ground-truth quirks.} D11's Metformin frequency is
  ``0-0-1 (BD)'' as written, which contradicts itself, so it
  is not scored. D10 carries a phone number by design, which
  probes number handling.
\item
  \textbf{Gates not recalibrated.} Tesseract's gate
  (\ensuremath{\tau} = 0.55) and PaddleOCR's gate
  (\ensuremath{\tau} = 0.87) were calibrated on axis-aligned
  OCR. We did not recalibrate either. For the rotated-crop
  and replacement pipelines we kept \ensuremath{\tau} =
  0.87, and Section 5.8 shows 21 readable handwritten slots
  on captures it rejects.
\end{itemize}

\hypertarget{statistical}{%
\subsection{7.5 Statistical}\label{statistical}}

With 15 documents, the document-level bootstrap has few
clusters, and its intervals are themselves approximate and
wide. Several intervals for the improvements have lower
bounds close to zero, so they are fragile. We made many
comparisons (seven planned configurations, four post-hoc
ones, a three-step ablation, several subsets) and did not
correct for multiplicity. Several intervals have lower
bounds near zero (+1, +2, +3, +5 points). Generation was
verified deterministic for the 29 October pipeline (Section
5.1), so there is no sampling variance there. For the
replacement pipeline we do not report that check, and each
cell is one run.

\hypertarget{conduct-and-reproducibility}{%
\subsection{7.6 Conduct and
reproducibility}\label{conduct-and-reproducibility}}

One author designed the pipeline, wrote the handwriting,
captured the photographs, designed the post-hoc prototypes,
and ran the evaluation. No independent party replicated any
step. The replay harness is debug-only and was introduced
alongside the evaluation, though production output was
byte-identical across all four builds (Section 4.4). The
rotated-crop, line-scope and tolerant-grounding changes of
Section 5.7 were verified by replay on the device, not in
the live camera flow. The replacement pipeline's final
backstop step was applied to logged model output in a
desktop harness, not on the phone (Section 4.4). The code,
the vision-language model's integration and the capture
photographs (about 820 MB) are not currently released, so
others cannot yet reproduce the replays. We describe the
pipeline in Section 3, the configurations in Section 4.5,
and give the test documents in Appendix A.

\hypertarget{what-the-study-does-not-show}{%
\subsection{7.7 What the study does not
show}\label{what-the-study-does-not-show}}

It does not show that the pipeline is accurate, safe, or
clinically useful, that any configuration generalises, that
a 2B vision-language model reads handwriting from other
writers, that the larger model is better as a model once its
template is held fixed, or that handwriting can be read at a
given accuracy. It shows where the 29 October pipeline loses
medications on this set, that a small deterministic backstop
kept false outputs from reaching the user at no cost in
recall, and what a change of crops, model and grounding does
on the same captures, and at what cost on the phone. Each of
these needs data that does not come from the same author.

\hypertarget{conclusion}{%
\section{8 Conclusion}\label{conclusion}}

We replayed identical frozen inputs through a phone-based
prescription reader and compared configurations of it, with
deterministic generation and paired document-level
intervals. The deterministic backstop removed no correct
medication and cut unsafe outputs from 45 to 1. Production
recall was low, 8\% overall and 0 of 45 on one writer's
handwriting. The causes differed by input: on handwriting
the text rarely reached the model in usable form, and on
typed sheets the model chose the wrong text. A scoped
prompt, a larger model and a switched-off repetition penalty
each raised recall on typed text but not on handwriting, and
numeric grounding caused most dosing errors. We also
withdrew an earlier claim that better OCR lowered recall,
which did not survive a controlled comparison. An
exploratory analysis traced much of the handwriting floor to
a defect in our own crop step. We then replaced the pipeline
with rotated-box crops, a 2B-parameter vision-language model
that re-reads and extracts, and per-line numeric grounding.
On the same captures recall rose from 8\% to 67\%, typed
recall from 19\% to 100\% and handwritten recall from 0 of
45 to 20 of 45, with no unsafe output. Rotated crops did
most of the work on reading accuracy and the model most of
the work on medications found, and the OCR confidence gate
is now the main limit. The price is about a minute per
photograph and about 1.9 GB of memory. All results are
in-sample, from one writer, 15 synthetic documents and one
device, so they locate failures and measure one replacement;
they do not estimate performance on real prescriptions. We
give the test documents and ground truth in an appendix so
that others can build comparable sets and test whether the
findings hold elsewhere.

\hypertarget{references}{%
\section{References}\label{references}}

\begin{itemize}
\tightlist
\item
  Cui, C., Sun, T., Lin, M., Gao, T., et al.~(2025).
  \emph{PaddleOCR 3.0 Technical Report.} arXiv:2507.05595.
\item
  Devi M, S. and Prakash K, H. (2026). An End-to-End
  Pipeline for Handwritten Prescription Understanding:
  OCR--NLP Integration and Error Propagation Analysis.
  \emph{Shanlax International Journal of Arts, Science and
  Humanities} 13(S2-i4-Jan):123--133.
  doi:10.34293/sijash.v13iS2-i4-Jan.10593.
\item
  Efron, B. and Tibshirani, R. J. (1993). \emph{An
  Introduction to the Bootstrap.} Chapman \& Hall, New York.
\item
  Field, C. A. and Welsh, A. H. (2007). Bootstrapping
  clustered data. \emph{Journal of the Royal Statistical
  Society, Series B} 69(3):369--390.
\item
  Geifman, Y. and El-Yaniv, R. (2017). \emph{Selective
  Classification for Deep Neural Networks.}
  arXiv:1705.08500.
\item
  Gemma Team (2025). \emph{Gemma 3 Technical Report.}
  arXiv:2503.19786.
\item
  ggml-org. \emph{llama.cpp: LLM inference in C/C++.}
  Software. \url{https://github.com/ggml-org/llama.cpp}.
\item
  Google (2025). Introducing Gemma 3 270M: The compact model
  for hyper-efficient AI. Google Developers Blog, August
  2025.
  \url{https://developers.googleblog.com/en/introducing-gemma-3-270m/}.
\item
  Henry, S., Buchan, K., Filannino, M., Stubbs, A. and
  Uzuner, O. (2020). 2018 n2c2 shared task on adverse drug
  events and medication extraction in electronic health
  records. \emph{Journal of the American Medical Informatics
  Association} 27(1):3--12. doi:10.1093/jamia/ocz166.
\item
  Islam, M. M. (2026). \emph{RxHandBD: A Handwritten
  Prescription Word Image Dataset.} Mendeley Data, v3.
  doi:10.17632/dsb5r6vskg.3.
\item
  Jagatap, A., Merugu, S. and Comar, P. M. (2025). RxLens:
  Multi-Agent LLM-powered Scan and Order for Pharmacy. In
  \emph{Proceedings of NAACL 2025, Industry Track},
  pp.~822--832.
\item
  Jayaraman, I. S. (2026). \emph{AgentOS.} Zenodo.
  doi:10.5281/zenodo.22655167.
\item
  Ji, Z., Lee, N., Frieske, R., Yu, T., Su, D., Xu, Y.,
  Ishii, E., Bang, Y., Chen, D., Dai, W., Chan, H. S.,
  Madotto, A. and Fung, P. (2022). \emph{Survey of
  Hallucination in Natural Language Generation.}
  arXiv:2202.03629. doi:10.1145/3571730.
\item
  Keskar, N. S., McCann, B., Varshney, L. R., Xiong, C. and
  Socher, R. (2019). \emph{CTRL: A Conditional Transformer
  Language Model for Controllable Generation.}
  arXiv:1909.05858.
\item
  Li, M., Lv, T., Chen, J., Cui, L., Lu, Y., Florencio, D.,
  Zhang, C., Li, Z. and Wei, F. (2021). \emph{TrOCR:
  Transformer-based Optical Character Recognition with
  Pre-trained Models.} arXiv:2109.10282.
\item
  Mankash, T., Kota, V. S. C., De, A., Prakash, P. and
  Jadhav, K. (2024). \emph{MIRAGE: Multimodal Identification
  and Recognition of Annotations in Indian General
  Prescriptions.} arXiv:2410.09729.
\item
  Rahim, A. (2026). \emph{Doctor's Handwritten Prescription
  BD dataset.} Mendeley Data, v2.
  \url{https://data.mendeley.com/datasets/zjjtptvn6f}.
\item
  Smith, R. (2007). An Overview of the Tesseract OCR Engine.
  In \emph{Proceedings of the Ninth International Conference
  on Document Analysis and Recognition (ICDAR)},
  pp.~629--633.
\item
  Ugare, S., Suresh, T., Kang, H., Misailovic, S. and Singh,
  G. (2024). \emph{SynCode: LLM Generation with Grammar
  Augmentation.} arXiv:2403.01632.
\item
  Wang, P., Bai, S., Tan, S., Wang, S., Fan, Z., Bai, J.,
  Chen, K., Liu, X., Wang, J., Ge, W., Fan, Y., Dang, K.,
  Du, M., Ren, X., Men, R., Liu, D., Zhou, C., Zhou, J. and
  Lin, J. (2024). \emph{Qwen2-VL: Enhancing Vision-Language
  Model's Perception of the World at Any Resolution.}
  arXiv:2409.12191.
\item
  Yang, A., Li, A., Yang, B., et al.~(2025). \emph{Qwen3
  Technical Report.} arXiv:2505.09388.
\end{itemize}

\hypertarget{appendix-a.-test-documents-and-ground-truth}{%
\section{Appendix A. Test documents and ground
truth}\label{appendix-a.-test-documents-and-ground-truth}}

Medication lines as written on the sheets (dose, frequency
code, duration). Patient, doctor and clinic names,
registration numbers and phone numbers are omitted. D01 to
D06 are typed, D07 to D12 are handwritten at tier
\emph{average}, and D13 to D15 at tier \emph{poor}.

\begin{longtable}[]{@{}
  >{\raggedright\arraybackslash}p{(\columnwidth - 4\tabcolsep) * \real{0.0575}}
  >{\raggedright\arraybackslash}p{(\columnwidth - 4\tabcolsep) * \real{0.3563}}
  >{\raggedright\arraybackslash}p{(\columnwidth - 4\tabcolsep) * \real{0.5862}}@{}}
\toprule\noalign{}
\begin{minipage}[b]{\linewidth}\raggedright
ID
\end{minipage} & \begin{minipage}[b]{\linewidth}\raggedright
Diagnosis
\end{minipage} & \begin{minipage}[b]{\linewidth}\raggedright
Ground-truth medications
\end{minipage} \\
\midrule\noalign{}
\endhead
\bottomrule\noalign{}
\endlastfoot
D01 & Type 2 diabetes mellitus, hyperlipidemia & Tab.
Metformin 500 mg, 1-0-1 (BD), 90 days; Tab. Atorvastatin 10
mg, 0-0-1 (OD), 90 days \\
D02 & Acute gastritis & Cap. Pantoprazole 40 mg, 1-0-0 (OD),
14 days \\
D03 & Upper respiratory tract infection & Tab. Azithromycin
500 mg, 1-0-0 (OD), 3 days; Tab. Cetirizine 10 mg, 0-0-1
(OD), 5 days \\
D04 & Essential hypertension & Tab. Amlodipine 5 mg, 0-0-1
(OD), 90 days; Tab. Telmisartan 40 mg, 1-0-0 (OD), 90
days \\
D05 & Viral fever & Tab. Paracetamol 650 mg, 1-0-1 (BD), 5
days \\
D06 & Allergic rhinitis & Tab. Montelukast/Levocetirizine 10
mg, 0-0-1 (OD), 15 days \\
D07 & Type 2 diabetes mellitus & Tab. Glimepiride 2 mg,
1-0-0 (OD), 30 days \\
D08 & Osteoarthritis (knee) & Tab. Shelcal 500 mg, 1-0-0
(OD), 60 days; Tab. Aceclofenac 100 mg, 1-0-1 (BD), 10
days \\
D09 & Acid reflux; skin infection (cellulitis, left leg) &
Cap. Pantoprazole 40 mg, 1-0-0 (OD), 14 days; Cap.
Amoxicillin 500 mg, 1-1-1 (TDS), 7 days \\
D10 & Common cold & Tab. Cetirizine 10 mg, 0-0-1 (OD), 5
days \\
D11 & Hypertension, type 2 diabetes, dyslipidemia & Tab.
AVAS 10 mg, 0-0-1 (OD), 90 days; Tab. Metformin 500 mg,
1-0-1 (BD), 90 days; Tab. Losartan 50 mg, 1-0-0 (OD), 90
days \\
D12 & Migraine & Tab. Naproxen 250 mg, 1-0-1 (BD), 5 days \\
D13 & Upper respiratory tract infection & Tab. Azithromycin
500 mg, 1-0-0 (OD), 3 days \\
D14 & Urinary tract infection & Cap. Nitrofurantoin 100 mg,
1-0-1 (BD), 5 days \\
D15 & Diabetes, hyperlipidemia, hypertension & Tab.
Metformin 500 mg, 1-0-1 (BD), 90 days; Tab. AVAS 10 mg,
0-0-1 (OD), 90 days; Tab. Amlodipine 5 mg, 0-0-1 (OD), 90
days \\
\end{longtable}

D11's Metformin frequency was written as ``0-0-1 (BD)'' on
the handwritten sheet, which contradicts itself, so its
frequency is not scored. D10 carries a phone number on the
sheet. Some sheets also carry a patient name, age and sex,
vitals, an advice line and a signature, which are not part
of the ground truth.
\end{document}